\documentclass[final]{nesy2025} % Anonymized submission for Openreview
\usepackage{amsmath,amsfonts,bm}

\def\eqref#1{equation~\ref{#1}}
\def\1{\bm{1}}

\DeclareMathAlphabet{\mathsfit}{\encodingdefault}{\sfdefault}{m}{sl}
\SetMathAlphabet{\mathsfit}{bold}{\encodingdefault}{\sfdefault}{bx}{n}

 \usepackage{url}

 \usepackage{graphicx}
 \graphicspath{{figures}}
 
 \usepackage{color}
 
 \usepackage{amsmath} % For math symbols like \mathrm \textsuperscript{}
 \usepackage{booktabs}
 \usepackage{multirow}
 \usepackage{rotating}
 \usepackage{array}
 \usepackage{booktabs}
 \usepackage{float}
 \usepackage{amssymb}
 \usepackage{svg}

\usepackage[disable]{todonotes}
\setuptodonotes{inline} % put all inline

 \usepackage{enumitem}
 \setlist{nosep}
 
 \PassOptionsToPackage{hyphens}{url}\usepackage{hyperref}
 \usepackage{cleveref}
 
 \usepackage[framemethod=tikz]{mdframed} % Used for highlighting entire sections
 \usepackage{soul} % Used for strikethroughs

\title[Do GNN States Contain Graph Properties?]{Do Graph Neural Network States Contain Graph Properties?
}

  \author{\Name{Tom Pelletreau-Duris} \Email{t.a.p.pelletreau-duris@vu.nl}\\
   \Name{Ruud van Bakel} \Email{r.van.bakel@vu.nl}\\
   \Name{Michael Cochez} \Email{m.cochez@vu.nl}\\
   \addr Vrije Universiteit, Amsterdam}

\begin{document}

\maketitle

\begin{abstract}
% \todo[inline]{Update abstract on openreview}
Deep neural networks (DNNs) achieve state-of-the-art performance on many
tasks, but this often requires increasingly larger model sizes, which in turn leads
to more complex internal representations. Explainability techniques (XAI) have made remarkable progress in the interpretability of ML models. However, the non-euclidean nature of Graph Neural Networks (GNNs) makes it difficult to reuse already existing XAI methods. While other works have focused on instance-based explanation methods for GNNs, very few have investigated model-based methods and, to our knowledge, none have tried to probe the embedding of the GNNs for structural graph properties. In this paper we present a model agnostic explainability pipeline for Graph Neural Networks (GNNs) employing diagnostic classifiers. We propose to consider graph-theoretic properties as the features of choice for studying the emergence of representations in GNNs. This pipeline aims to probe and interpret the learned representations in GNNs across various architectures and datasets, refining our understanding and trust in these models.
\end{abstract}

\section{Introduction}

Graph Neural Networks (GNNs) are pivotal in harnessing 
%non-Euclidean 
graph-structured data \citep{kipf2017semi-supervised} for tasks ranging from social network analysis to bioinformatics. Despite their success, the black-box nature of GNNs poses significant challenges as classical XAI methods cannot be directly applied on GNNs due to the lack of a regular structure (e.g. vertices can have different degrees). 
In this case, explaining a prediction means identifying important parts of the relational structure, or input features of nodes. An issue is that finding the explanation is itself a combinatorial problem, making  XAI methods for GNN intractable \citep{ying2019gnnexplainer,longanodateunderstanding}.

%For node prediction for example, it involves identifying the right relational structure based on the node of interest that explains the prediction. Explaining a graph prediction typically leads to the combinatorial complexity of the possible graph schemes that could explain the prediction. This complexity often makes XAI (explainable AI) methods for GNN intractable \citep{longanodateunderstanding,ying2019gnnexplainer,lucic2022-1cf-gnnexplainer}.

%Existing post-hoc GNN explanations methods \citep{agarwal2023evaluating,dai2022comprehensive} can be classified into two main categories: \textit{instance-level} and \textit{model-level} methods \citep{barredo_arrieta2020explainable}.

% maybe to put away who knows ? 
% In the realm of instance based methods, \textit{gradient-based} methods use the gradients of the output with respect to the input or intermediate features to measure the importance of each component of the graph. \textit{Decomposition-based} methods try to decompose the input graph into smaller subgraphs or paths that can account for the output. \textit{Surrogate-based} methods use a simpler, more interpretable model to approximate the behaviour of the original GNN and provide explanations based on the surrogate model.
% \todo[inline]{Writing feedback: when listing multiple references, if the order is not significant, order then from oldesrt to newest}
The surveys by \citet{dai2022comprehensive} and \citet{agarwal2023evaluating} highlighted the lack of comprehensive, robust and model-agnostic explainability methods. We also identified (see \cref{relatedwork}) few model-level explainability methods. Most works in GNN explainability present methods that identify a set of nodes and possibly their attributes as the explanation of a prediction \citep{vu2020pgm-explainer,zhang2021relex,saha2022model-centric,azzolin2023globalexplainabilitygnnslogic,wang2023reinforced,xuanyuan2023globalconceptbasedinterpretabilitygraph}. 
% \todo[inline]{Writing feedback: do not use contractions like don't, isn't, it's, ... in academic writing.}
Other works focus on explaining the decision-making process at a high level, often by generating graph patterns or motifs that influence the predictions \citep{ying2019gnnexplainer,yuan2020xgnn}. 
However, oftentimes there is more information in the structure of the relationships between elements than in the juxtaposition of the elements themselves. 
Graph prediction ought to take into account structure 
% (a cluster, motifs across clusters, or higher order properties across the graph) 
that is invariant across the graph hierarchies and symmetries. 
Achieving this, we also get interpretability of intermediate layers, which previous methods do not provide. 

\begin{figure}[h]
    \centering
    \includegraphics[width=\textwidth]{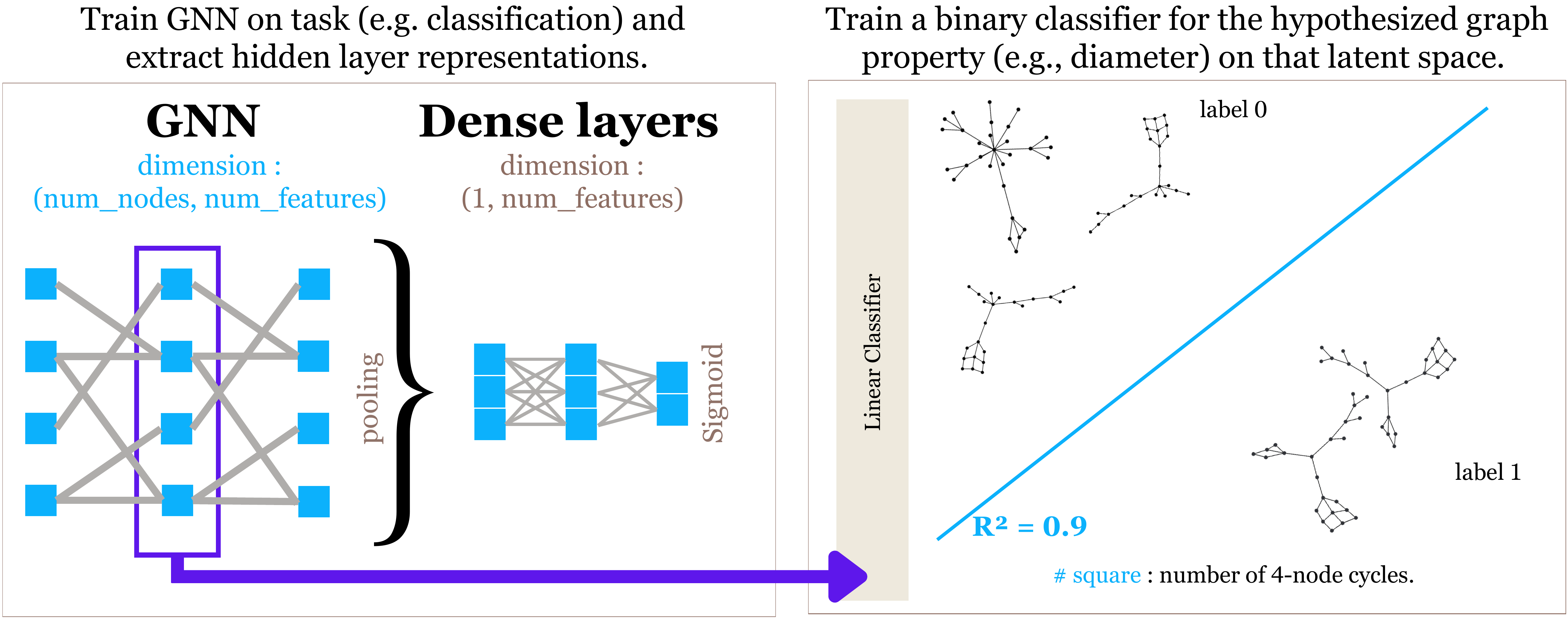}
    \caption{Illustration of the probing pipeline. Note that graphs of label 0 have only grid or house while those of label 1 have both. If a linear probe has good performance ($R^2$ score) then there exists a hyperplane in the representation space that separates the inputs based on the property.}
    \label{fig:probing_GNN_Pipeline}
\end{figure}

One prior work identified the role probing classifiers could play \citep{akhondzadeh_probing_2023}, as developed for Natural Language Processing \citep{giulianelli2018under,belinkov2021probing}. 
That work focused on whether the hidden representation encoded the number of hydrogen atoms or the presence of aromatic rings. 
We aim to address the more fundamental and general question; finding out whether the hidden representation encodes information about the graph-theoretic properties. 
We propose graph theoretic properties as the best candidate for studying GNN inductive bias. 
Graph theoretic 
% \todo[inline]{MC: consistency; either we call them properties, or features, no mixing.}
properties are natural algorithmic abstractions of graphs used in network science \citep{BARABASI2002590}. 
They would act as representational atoms \citep{bereska_mechanistic_2024}, as features that GNNs would leverage to render graph classification problems linearly separable. The main hypothesis tested in this paper is whether GNNs leverage such algorithmic strategies to preserve, abstract, or discard structural information through the network’s layers. 
This work bridges GNN interpretability with the broader theory of linear feature representations in neural models \citep{nanda2023emergentlinearrepresentationsworld,bereska_mechanistic_2024}. We propose a model-agnostic pipeline to interpret GNN embeddings by probing for encoded graph properties (see \cref{fig:probing_GNN_Pipeline}) across various architectures and datasets. We investigate both local properties like betweennes centrality, as well as global properties like average path length.
\footnote{ All code is publicly available from \\
%\phantom{blablablablablablablablablablablablablablablablablablablablablablablablablablablablabla}
\url{https://github.com/TomPelletreauDuris/Probing-GNN-representations}}

%IMPORTANT OR NOT ?
% The probing pipeline is illustrated in \cref{fig:probing_GNN_Pipeline}. First, a GNN is trained on a specific task, such as classifying between smaller graphs (label 0) and larger graphs (label 1). Similar to how a CNN might organize images of similar shapes or textures into distinct regions in its feature space, the GNN embeddings might arrange graphs based on structural properties like their diameter. Next, we extract embeddings from the internal layers of the GNN. These embeddings are used to train the probing model—in this case, a binary classifier tasked with determining whether the embeddings encode predictive information about the graph's diameter. If a linear probe has good performance ($R^2$ score) then there exists a hyperplane in the representation space that separates the inputs based on the property.

Our core contribution is that we show that using a diagnostic classifier, as illustrated in \cref{fig:probing_GNN_Pipeline}, we can effectively highlight graph-theoretic properties in GNN learned latent representations (\cref{fig:plotGCNTSNEcontrol}). We further explore how different regularization techniques 
%(none, $L_2$ weight decay, dropout) 
affect the representation of graph properties (see \cref{fig:plotGINL2}). 
We also investigate how the GNN architecture (\cref{tab:gridhousegraphprobing}) and datasets (\cref{table:Clintoxperformance}) affect the probing. 

\section{Background}

% \todo[inline]{MC: In this section, we can probably compress/cut a lot. }

% mathematical restictions of GNNS
\subsection{Graph Neural Networks}

\textbf{Graph Convolutional Network} (GCN)~\citep{kipf2017semi-supervised} are GNNs where for a single layer, the node representation is computed as:
$
\boldsymbol{X}^{\prime}=\sigma\left(\tilde{\boldsymbol{D}}^{-1 / 2} \cdot \tilde{\boldsymbol{A}} \cdot \tilde{\boldsymbol{D}}^{-1 / 2} \cdot \boldsymbol{X} \cdot \boldsymbol{W}\right)
$.
We know that GNNs which rely solely on local information, like the \textbf{GCN} and its relational variant (\textbf{R-GCN})~\citep{schlichtkrull2018modeling}, cannot compute important graph properties, such as girth and diameter or eigenvector centrality~\citep{garg2020generalizationrepresentationallimitsgraph}. 
We are therefore also investigating more globally aware networks 
%and Weisfeiler-Lehman tested one like %\textbf{k-hop GCN}~\citep{nikolentzos2020khopgraphneuralnetworks,feng2023powerfulkhopmessagepassing}\todo{Did we do the k-hop GCN in the end?}, 
like \textbf{GAT} (Graph Attention Network)~\citep{velickovic2018graph} and \textbf{GIN} (Graph Isomorphism Network)~\citep{xu2019how}. The models expressivity is based on the Weisfeiler-Lehman test~\citep{akhondzadeh_probing_2023}. 
GIN aggregates node features in a way that mimics the Weisfeiler-Lehman test. By using the MLP equivalent to an Injective Update Function, GIN avoids oversimplifying the aggregation step, making it as expressive as the WL test. Thus, it is likely to excel at encoding complex graph properties and solving classification tasks. 
%Self attention used in 
GAT should theoretically 
%make them 
be as expressive while we expect GAT to be slightly less expressive, GCN even less. 
% GAT makes use of self-attention and is thereby more expressive than the GCN. However, its reliance on feature-dependent weights and structure-free normalisation limits its ability to capture specific structural properties that do not directly depend on edges. 
% This is particularly true for tasks where node features alone are not enough, and global graph structures are crucial (e.g., tasks requiring knowledge of subgraphs or non-local patterns). 

\subsection{Graph properties}
Graph theory is a branch of mathematics that studies the properties and relationships of graphs. Graphs can be undirected or directed and analysed through both local and global properties. Local properties (like node degree or clustering coefficient) are based on the neighbors of a node.
In contrast, global properties (such as diameter and characteristic path length) assess the overall graph structure. Global graph properties can be associated with higher level complex systems' characteristics like the presence of repeated motifs in the graphs or information-flow properties.
See the \cref{sec:appendix:graph_properties} for a list of local and global properties used in our experiments.
We can distinguish different global properties, \textit{basic} ones like the number of nodes a graph has, \textit{clustering and centrality} ones, \textit{graph motifs and substructures}, \textit{spectral and small-world properties}. 
As an higher-order analysis, the recurrence of specific motifs within network substructures—such as triangles, cliques, or feed-forward loops can be seen as the fundamental building blocks that dictate the system's functionality and resilience.
Small-worldness, as characterised by Barabási \citep{barabasi1.74.47}, reveal how networks can maintain short path lengths despite their expansive size and sparse connectivity. 
% This kind of higher order properties are very interesting in order to understand how the macroscopic behaviour of complex systems emerges from the intricate interplay of their microscopic components \citep{BARABASI2002590}. For example how diseases spread in social networks, how neurons interact in the brain, or how information propagates through the Internet. 
GNNs synthesise local topological features into global structures, abstract these representations into higher-order graph attributes. Each layer progressively expands the receptive field, aggregating local neighborhood properties that are relevant for the classification towards high level graph properties, mirroring how hierarchical feature learning works in convolutional neural networks (CNNs) for images. 
% Probing their learnt representations should act as a scalable proxy to investigate how global arrangement and connectivity patterns influence a system's function. In other terms, by dissecting these learned embeddings, we can possibly delve into the intricate relationships between a network's macroscopic arrangement and its emergent behaviours.
Through hierarchical pooling or readout mechanisms, GNNs can aggregate node embeddings into a single, global graph-level embedding. Based on the message passing paradigm in GNNs, as layers progress, one would expect an increased abstraction in the selection of graph properties. Initially, local features like node degree should dominate, but deeper layers progressively should capture more global properties, such as connectivity patterns and centrality. 
% Graphs that share structural similarities or patterns of interaction among nodes are organised closely in the embedding space, allowing the model to differentiate between classes of graphs, such as those with and without long paths (\cref{fig:probing_GNN_Pipeline}). This is why, AI engineers are likely to focus on the validation framework, emphasizing the alignment between probing results and the predictions derived from the message-passing paradigm—particularly insights revealed in the initial layers of the model. In contrast, from the standpoint of domain researchers—such as chemists or neuroscientists— the most compelling aspect is found in the later layers of the model, where the abstract representations become increasingly capable of rendering the problem linearly separable, thus facilitating clear interpretability of classification decisions and offering domain-specific interpretations.
Graph-theoretic properties serve a symbolic role, offering interpretable, human-defined structure, while emergent features in GNNs reflect a connectionist process, arising through distributed representations shaped by task supervision.
% from an AI engineer's perspective, the primary focus would lie on the first-layer representations, where the key question is whether they align with the expected behavior of the message-passing paradigm or self-attention mechanisms. In contrast, from the standpoint of domain researchers—such as chemists or neuroscientists—the most compelling aspect is found in the later layers of the model, where the abstract representations become increasingly capable of rendering the problem linearly separable, thus facilitating clear interpretability of classification decisions.

\subsection{Probing classifiers}

In prior work~\citep{hupkes2018visualisation} probing classifiers have been used for linguistic properties. The probing paradigm is a post-hoc explainability method \cite{alain2018understanding,belinkov2021probing}. Probing is a concept-based approach in the larger field of mechanistic interpretability \citep{bereska_mechanistic_2024}. Probing the learned representations for high-level concepts and patterns is a top-down approach to unravel a model's decision-making process. It's also called representation engineering \citep{zou2025representationengineeringtopdownapproach}.
Here, we adapt them for graph features. Unlike unsupervised techniques such as PCA or T-SNE, which are useful to visualize input data with regard to the embedding latent space, we adopt a supervised framework to quantitatively assess how specific properties are encoded within the embedding space of GNNs. 
Let \( g : f_l(x) \mapsto \hat{z} \) represent a probing classifier, used to map the learned intermediate representations from the original model \( f \) to a specific property \( \hat{z} \). The choice of a linear classifier for \( g \) is motivated primarily by its simplicity. If a linear probe performs well, it suggests the existence of a hyperplane in the representation space that separates the inputs based on their properties, indicating linear separability.

The justification for simple linear probe comes from the linear representation hypothesis \citep{neal2017how}. The linear representation hypothesis proposes that features are directions in activation space, i.e. linear combinations of neurons \citep{bereska_mechanistic_2024}. It's well studied in LLMs \citep{park_geometry_nodate}. Rather than assuming this hypothesis holds, linear probes can be taken as diagnostic tools to investigate whether learned representations organize information in linearly accessible ways for particular concepts and datasets.

Another advantage of a simple linear probe is avoiding the risk that a more complex classifier might infer features that are not actually used by the network itself \citep{hupkes2018visualisation}. While other non-linear probes have been explored in the literature \citep{belinkov2021probing}, even studies showing improved performance with complex probes maintain the same logic: \( \operatorname{Perf}(g, f_1, \mathcal{D}_O, \mathcal{D}_P) > \operatorname{Perf}(g, f_2, \mathcal{D}_O, \mathcal{D}_P) \) holds across representations \( f_1(x) \) and \( f_2(x) \) when evaluated by a consistent probe \( g \). This consistency ensures valid comparison, underscoring that if a property can be predicted well by a simple probe, it is likely relevant to the primary classification task. 
% From an information-theoretic perspective, training the probing classifier \( g \) can be viewed as estimating the mutual information between the learned representations \( f_{l}(x) \) and the property \( z \). This mutual information is denoted as \( \mathrm{I}(\mathbf{z}; \mathbf{h}) \), where \( \mathbf{z} \) refers to the property and \( \mathbf{h} \) represents the intermediate representations \citep{belinkov2021probing}.

Mathematically speaking, this supervised approach allows us to define hyperplanes or higher-dimensional decision boundaries that partition the embedding space according to the chosen graph property. The \( R^2 \) score serves as this information-theoretic measure indicating how well the hyperplane divides the inputs in the embedding space. The $R^2$
score (see \cref{R2score} for a formal definition) measures the proportion of variance in the dependent variable that is predictable from the independent variable(s). A \( R^2 \) near 1 indicates that the embeddings are highly informative about \( \hat{z} \), suggesting that the neural model has internalized this property in a linearly accessible manner.

By defining specific properties that could divide the embedding space and assessing how well the corresponding hyperplanes make the embedding space linearly separable, we gain quantitative insights into the abstract features aggregated within the embeddings. 
This method moves beyond mere hypothesis generation based on clustering patterns observed through techniques like PCA, providing a rigorous framework for understanding how well the embedding space represents complex graph properties. It can also be thought as complementary from the T-SNE and PCA visualisation techniques. It provides a quantitative measure of the separability of the embeddings based on the property of interest. Our best empirical observation of this is the correspondence between the T-SNE visualisation of embeddings and their corresponding \( R^2 \) scores in \cref{fig:plotGINTSNEcontrol,fig:plotGCNTSNEcontrol,fig:plotGATTSNEcontrol}.

\section{Datasets}

% Our experiments are set up with three different datasets: the artificially created \textbf{Grid-House} dataset, a dataset containing molecules and their toxicity \textbf{ClinTox Molecular} and an fMRI brain scan dataset \textbf{fMRI FC connectomes} .

All three datasets have the same setup: given a set of graphs \(\left\{\mathcal{G}^1, \mathcal{G}^2, \ldots, \mathcal{G}^N\right\}\), predict the corresponding binary labels$\left\{y^1, y^2, \ldots, y^N\right\}$.

\begin{figure}[t]
\centering
\includegraphics[width=0.9\textwidth]{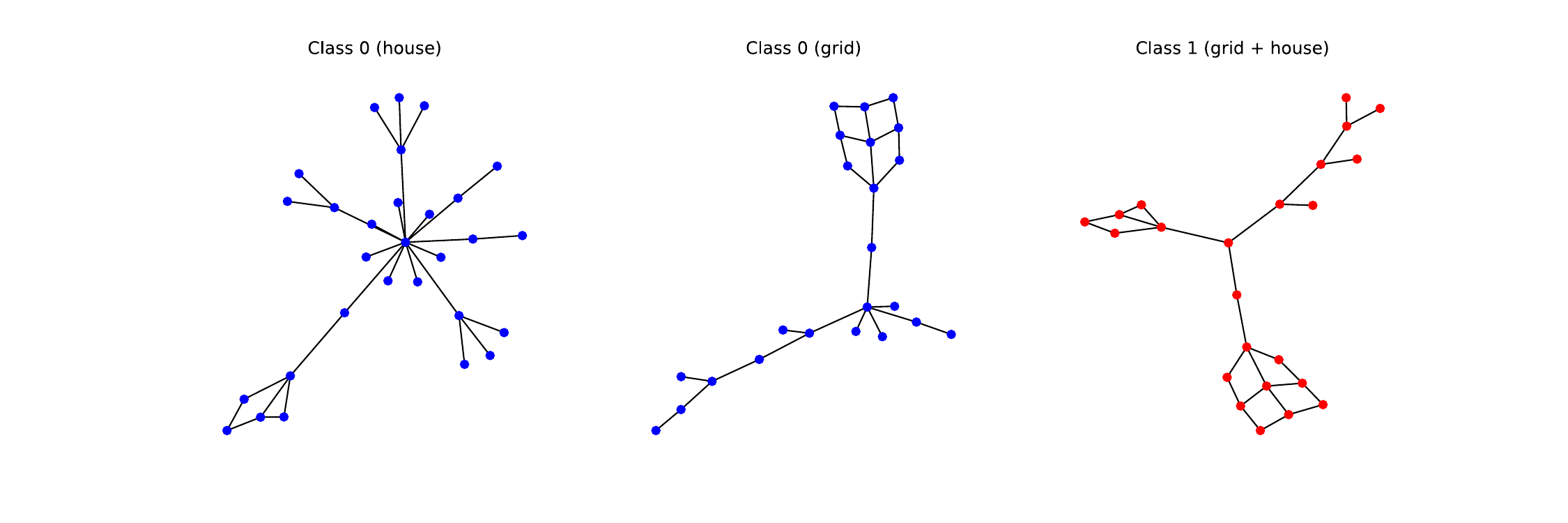}
\caption{Illustration of the grid-house dataset. The first class (0) include graphs with either a house (square+triangle) either a 3x3 grid (4 squares). The second class include both a house and a grid.} \label{fig2}
\label{fig:grid_house}
\end{figure}

\paragraph{The Grid-House dataset} inspired by \citep{agarwal2023evaluating} is designed to evaluate the compositionality of GNNs. It features two concepts: a 3x3 grid and a house-shaped graph made of five nodes. The dataset consists of Barabási-Albert (BA) graphs~\citep{barabasi2} with a normal distribution of the number of nodes. The negative class includes a BA graph connected to \textit{either} a grid or a house, while the positive class contains a BA graph connected to \textit{both} a grid and a house (see \cref{fig:grid_house}). For accurate classification, models need to identify and combine simple patterns. Recognizing isolated patterns or single node features is not sufficient. In order to ensure that the classification can't be solved by simply counting the number of nodes in a graph, the average number of nodes has to be the same between classes. The number of nodes is uniformly distributed between 6 and 21 for the grid graphs, between 7 and 22 for the house graphs, and between 1 and 16 when both are present. During generation, we ensure no test set leakage by removing isomorphisms. On 2,000 graphs, we apply an 80/20 train/test split.
% The dataset is randomly divided between the train (80\%) and the test set (20\%).
%This setup challenges models to identify and combine simple patterns, as recognizing isolated patterns is insufficient for accurate classification. 
%The recognition of a single node feature alone can't be sufficient and the recognition of a single motif alone neither. 
%The need of long-range dependency would work. 
The dataset helps investigate how GNNs combine multiple concepts and addresses the ``laziness'' phenomenon, where networks learn patterns characterising only one class and predict the other by default~\citep{longa2023explaining}.

%We also addressed concerns regarding potential limitations in the dataset generation process, particularly the possibility of data leakage between the training and test sets. Since the probability of leakage is directly related to the likelihood of generating isomorphic graphs, which depends on the inherent randomness in the graph generation process (including variations in the number of nodes, the Barabási–Albert graphs iterative modelling, the motifs attachment configurations), we estimated that the probability of generating isomorphic graphs was low but significative in a dataset of 2,000 samples. To ensure the integrity of the data splits and to eliminate any residual risk of data leakage, we opted to manually remove any isomorphic graphs from the dataset.

The dataset has been structured such that an optimal, linearly separable solution requires the identification of global structural motifs in the graph. A good strategy would be counting the number of squares (i.e., four-node cycles). Important note, a random Barabási-Albert graph cannot contain any four-node cycles. Meanwhile, a grid subgraph will consistently exhibit four such cycles, while a house subgraph contains exactly one four-node cycle and one three-node cycle. Therefore, a graph that contains both a grid and a house will have a total of five four-node cycles. The presence of a three-node cycle could help the diagnostic of one type of graph in the negative class but is not necessary nor sufficient for solving the classification problem. Conversely, counting the number of four-node cycles is necessary and sufficient. In our example in \cref{fig:plotGINTSNEcontrol} we see that a GIN architecture organizes the input graphs in its latent space layer by layer. At layer 5 (MLP1) it even differentiates grid-only from house-only graphs in the negative class (purple).
% Thus, distinguishing between the classes does not really necessitates leveraging centrality-based measures but only recognizing the presence of a specific number of four-node cycles, enabling the model to effectively differentiate between the positive and negative classes. Thus the interesting results of \cref{fig:plotGINTSNEcontrol}.

The Grid-House dataset serves a critical purpose in our study: it offers a controlled and well-defined environment for rigorously testing our hypotheses. By construction, we are sure the classification problem can be solved in an algorithmic way. The question becomes whether GNNs inductive bias converge to this algorithmic strategy. The simplicity of these controlled constraints allow us to verify whether the GNN operates as intended in a setting where extraneous factors are minimized. 
% If a model struggles with this dataset, it would likely underperform on real-world, more complex graphs, underscoring its diagnostic value. 

% figure showing the grid house classificaiton. 

\paragraph{ClinTox Molecular} contains molecular graphs representing compounds with binary labels indicating whether they are toxic or non-toxic. The dataset consists of 1,491 drug compounds with known chemical structures. Each molecule is represented as a graph where nodes correspond to atoms and edges to bonds, with node features representing atom types and edge features representing bond types. The task is to predict toxicity. % and is often use in machine learning.

\section{Methodology} 

%The difference is in the setting: the artificial dataset, ClinTox Molecular and fMRI FC connectomes respectively use altered BA graphs, molecular graphs and weighted brain networks. Their repsective prediction tasks are: grid or house vs. grid and house (see fig. \ref{fig:grid_house}), toxic vs. non-toxic and healthy vs. ASD/MDD.

We propose a pipeline to confirm that graph-theoretic properties are a good candidate for studying GNN inductive bias. For each of the three datasets, we compare different GNN architectures (GCN, GIN, or GAT). The first layers are followed by a pooling operation (mean-~\citep{kipf2017semi-supervised}, sum-\citep{xu2019how}, or max-pooling~\citep{hamilton2017inductive}), and then a number of dense layers. For the \textbf{Grid-House dataset} the full hyperparameter settings after optimization can be found in \cref{table:brainib_hyperparameters}. Because we can only probe on one model weights, we ran each model 20 times and took the one with the best accuracy. There is a correlation of 0.992 between the accuracy and the maximum probing score \cref{fig:correlation}. 

We also compare different regularization methods and their effect on GNN representations. 
% Explicit \textbf{L2 loss regularisation} adds a penalty term to the loss function proportional to the square of the magnitude of the weights. This encourages the network to keep the weights small, preventing any single weight from becoming too large. By keeping weights small, we would expect the embeddings to become smoother and less sensitive to noise or fluctuations in the input data. We expect this to make the graph embeddings more linearly separable for features of interest.
It is known that explicit \textbf{$L_2$ regularization} encourages the network to keep the weights small. We expect that this will make the embeddings less sensitive to fluctuations in the input data which would translate in later layers being more selective to graph properties, leading to more specific receptive fields.
\textbf{Dropout} randomly disables a fraction of the neurons during each training iteration which forces the network to learn redundant representations, as any neuron could be dropped out. 
These redundant representations might make it more difficult to linearly separate the graph embeddings. We expect later layers to distinguish less between graph properties. In other words, we expect more polysemanticity \citep{bereska_mechanistic_2024}. We plot only post-pooling layers for the sake of clarity.

As a second set of experiments, we use our build pipeline to probe GNNs on real-life datasets, wondering if what we find aligns with domain knowledge in chemistry.
For the \textbf{ClinTox Molecular} dataset, we ranged the number of layers from 4 to 6 and hidden dimensions from 64 to 256. The final model architectures were selected based on optimal performance on this dataset. We also did preliminary work on Knowledge Graphs and fMRI connectomes, which is available in the appendix.
Before probing, we first sort the embeddings in descending order based on their norms before concatenating \cref{fig:plotGINTSNEcontrol}. This ordering depends only on the inherent properties of the embeddings themselves, not on their original ordering in the graph. As such, it inherently respects permutation invariance because reordering the nodes does not affect their norms or the resulting sorted order. Conveniently, sorting in this way ensures that any padding zeros align at the end of the sequence, enabling learnable representations for graphs with varying node counts. 
% While sorting for permutation invariance is not widely discussed in the literature, it provides a practical solution by using the embeddings' properties to enforce consistent ordering across graphs. 

\section{Results} 

\subsection{Grid-House dataset}

%\phantom{.} % Adds an invisible character, creating vertical space
All tested models achieve a test accuracy of over 0.9, with the GIN without regularization or dropout reaching the highest score of 1.0 (see \cref{table:gridhouseperformance} in the appendix for more details). The probing results in \cref{fig:plotGINTSNEcontrol} demonstrate that the \textit{number of squares} consistently yields the highest \(R^2\) scores. This focus on \textit{\#squares}, effectively partitions the graphs into two classes: those with \(\#\text{squares} < 5\) (indicating either the grid or house alone) and those with \(\#\text{squares} = 5\) (indicating the presence of both substructures). We see the same results for the GAT model (\cref{fig:plotGATTSNEcontrol}), especially for the final layers and for other model variants (see the appendix). This confirms our initial hypothesis where the number of square is the property of interest to perform this classification. Interestingly, the \textit{MLP1} layer of the GIN model can effectively separate the graph embeddings using the \textit{\#triangles} feature.

\begin{figure}[t!]
    \centering
    \includegraphics[width=0.9\textwidth]{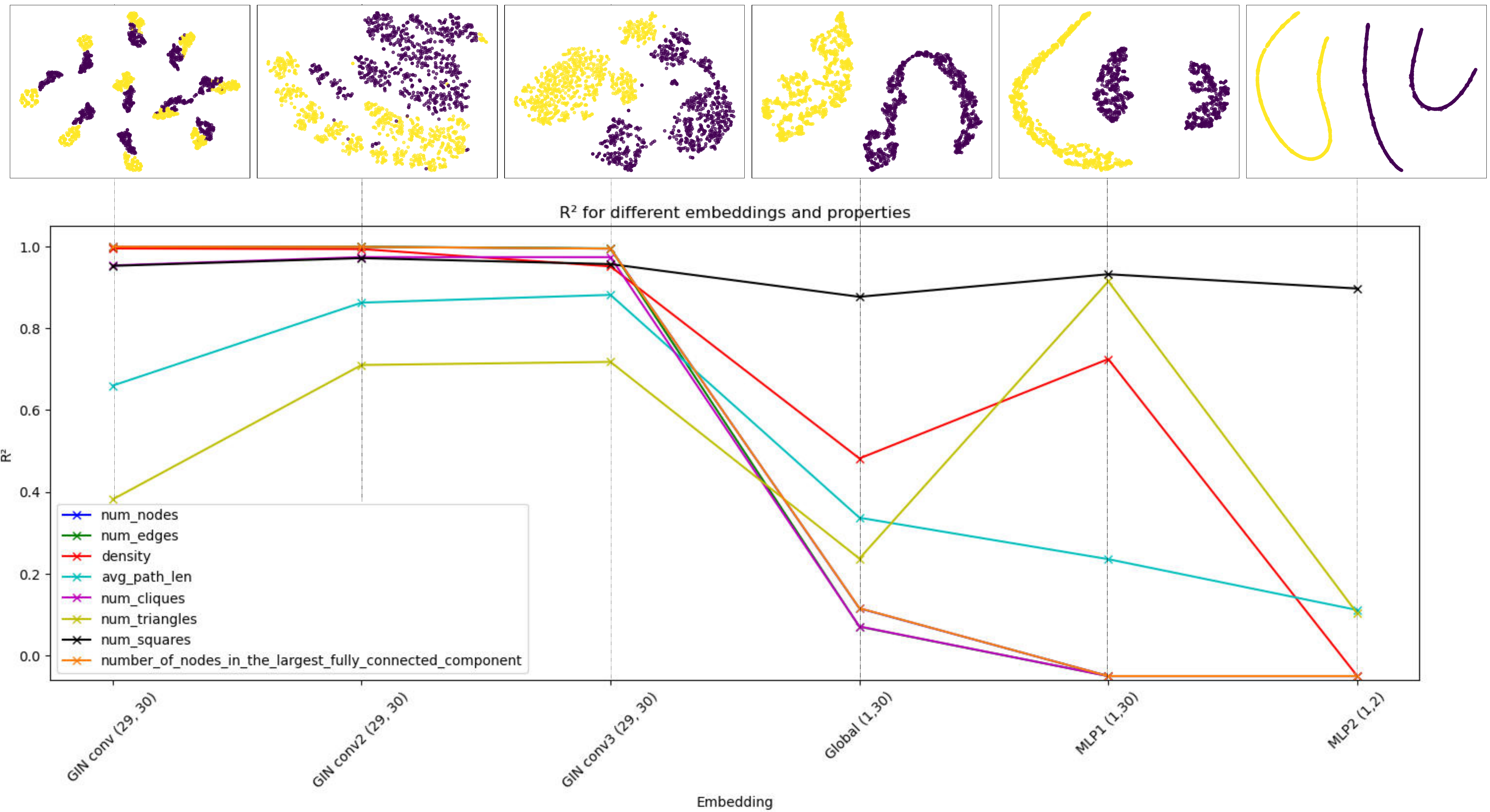}
    \caption{GIN probing results $R^2$ across different layers aligned with the T-SNE visualisations of the embedding (Grid House)}
    \label{fig:plotGINTSNEcontrol}
\end{figure}

\subsection{ClinTox Molecular}
As expected, the GIN model outperform the other models with a test accuracy of 0.93 (\cref{table:Clintoxperformance} in the appendix). We find that the highest performance is consistently achieved by probes for the features of \textit{average degree}, the \textit{spectral radius}, the \textit{algebraic connectivity} and the \textit{density}, in that order (see \cref{table:graph_properties_GIN}). 

\begin{figure}[t!]
    \centering
    \includegraphics[width=0.9\linewidth]{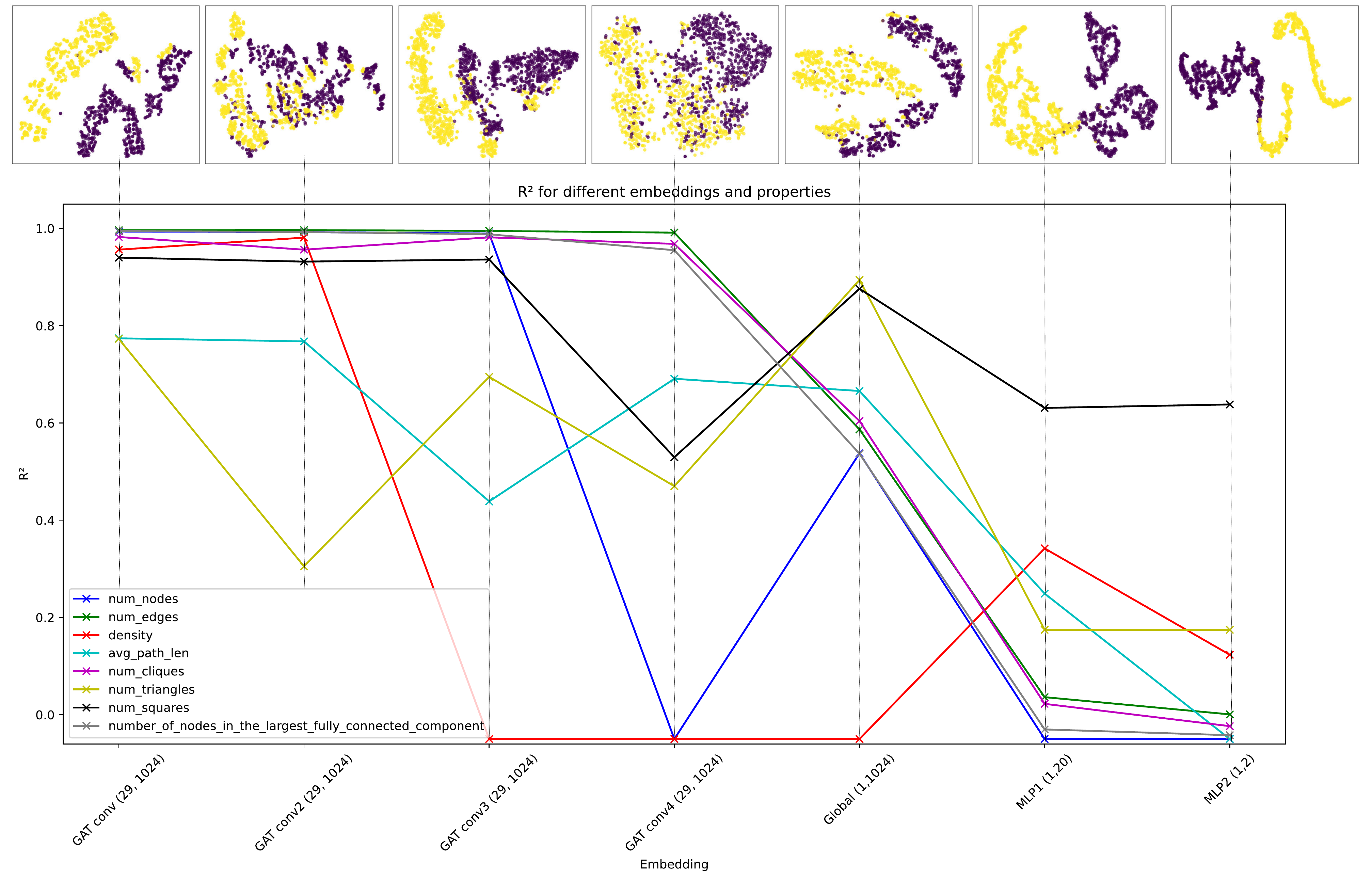}
    \caption{T-SNE visualization across different
layers of our GAT architecture aligned with the probing $R^2$ scores plots (Grid House)}
    \label{fig:plotGATTSNEcontrol}
\end{figure}

These findings validate our methodology on already known domain knowledge. Indeed, the average degree of atoms in a molecule provides a straightforward interpretation, as atoms with higher valencies are generally less stable and less biologically compatible. For instance, hydrogen with a valency of 1 and oxygen with a valency of 2 are more compatible with carbon-based molecules, whereas sulfur, with a valency of 6, is less favorable for biological systems ~\citep{komarnisky2003sulfur}. Therefore, the average degree serves as a useful indicator of molecular toxicity. Additionally, the spectral radius, often associated with molecular stability and reactivity, is another valuable graph property. Molecules with a lower spectral radius tend to be more stable, while those with a higher spectral radius may exhibit localized electron densities, increasing their reactivity. Domain knowledge on molecular toxicity align with the GIN strategy.

\subsection{Pooling}
While have deferred the results for regularization and dropout to the appendix, we did notice that regularization tends to help separate out the \#\text{squares} better, while dropout has the opposite effect. This is in line with our expectations.

\begin{table}[b!]
\centering
\caption{Linear Probing $R^2$ Performance Across GIN Layers for Selected Graph Properties (ClinTox Dataset). Best Scores in Bold; Non-convergence indicated by \textemdash}
\label{table:graph_properties_GIN}
\begin{tabular}{lcccccc}
\toprule
GIN Layer & Avg. degree & Sp. radius & Alg. co. & Density & Avg. btw. cent. & Graph energy \\
\midrule
x\_global & \textbf{0.81} & 0.74 & 0.67 & 0.58 & 0.48 & 0.44 \\
x6 (MLP)  & \textbf{0.80} & 0.74 & 0.66 & 0.58 & 0.42 & 0.44 \\
x7 (MLP)  & \textbf{0.75} & 0.71 & 0.56 & 0.50 & 0.47 & 0.46 \\
x8 (MLP)  & \textemdash & \textbf{0.07} & 0.02 & 0.00 & 0.06 & 0.05 \\
\bottomrule
\end{tabular}
\end{table}

\section{Discussion}
\textbf{Expectations:}
With \emph{Grid-House} we hypothesized that the GNNs would benefit from leveraging the \textit{\# of square} to render the problem linearly separable. Based on their mathematical restrictions, we hypothesized that the GIN would perform better than the GAT and the GCN. Regularization methods should either refine the representations, as seen with \( L_2 \) regularization, or distribute them more broadly, as achieved with dropout.
Based on the message-passing paradigm, we anticipated a clear absence of the \textit{\# of square} in the first layer. Additionally, we expected that the mean-pooling and norm-sorting methods would not significantly alter how representations are probed, except for basic properties like the number of nodes, which are easily interpretable from the tensor of node vectors but not from an aggregated representation.
For the \emph{ClinTox Molecular} dataset, based on the literature~\citep{chen2021chemical, jiang2021ggl,kengkanna2024enhancing} some properties have been found to be link with toxicity such as the node degree (i.e. the valency), subgraph patterns (functional groups, chemical fragments), and the overall graph connectivity.

% Based on existing literature on functional connectivity (FC) network properties in ASD and MDD \cref{brainimaginggnn}, we hypothesized that specific properties would be critical in classifying brain networks for the \textbf{fMRI FC connectomes} dataset. For ASD, we expect \textit{betweenness centrality} to play a significant role at the node level, reflecting local overconnectivity. At the graph level, we anticipate that \textit{clustering coefficient}, \textit{characteristic path length}, and \textit{small-worldness} will be essential in capturing the local and global network disruptions seen in ASD, particularly the imbalance between local overconnectivity and long-range underconnectivity. For MDD, we hypothesise that increased \textit{clustering coefficients}, \textit{modularity}, \textit{number of triangles} and \textit{number of squares} will be key features for classification, as they could indicate of heightened local interconnectedness and disrupted global integration.

\textbf{Findings:}
We first demonstrate the feasibility of our probing method through the \emph{Grid-House} dataset. In line with our expectations, The \textit{number of squares} metric dominated across all layers and models, with GIN showing enhanced expressivity. Secondary properties like \textit{avg\_path\_lenght} \cref{fig:plotGCNTSNEcontrol,fig:plotGCNTSNEcontrol_pooled} showed early significance but gradually diminished through the layers, demonstrating how GNNs act as low-pass filters on graph signals. The receptive field selects the most discriminative properties. These findings align with the labels; density or average path length increases when both a grid and a house are added to a graph. 
% Mean-pooling and norm-sorting methods exhibit different behaviors that can be explained by the type of information the classifier access : aggregated graph data or detailed node embeddings. 
% The probe might also infer relationships and patterns that are implicitly encoded in individual node embeddings but become explicit when considered collectively.

The inductive bias of Graph Neural Networks (GNNs), when viewed through the lens of the linear representation hypothesis, reflects a tendency to organize internal representations such that task-relevant features become increasingly linearly decodable as a function of the supervision signal. In this framework, the supervision signal guides the layer-wise development of representations, selectively reinforcing directions in activation space that align with informative graph-theoretic properties (e.g., centrality, motif counts, degree distributions). %GNNs internalize graph-theoretic structure in a supervision-aligned, geometrically linear fashion.

% Features useful for the prediction task are distilled into linear directions, while others may be attenuated or remain non-linearly entangled. This reflects not just the model’s architecture but also its task-driven inductive bias, shaped by the interplay between graph structure and supervision.
% We would expect unsupervised model to capture structural information more effectively than supervised ones \citep{kingma_auto-encoding_2022}. Unsupervised learning algorithms can discover hidden patterns and structures in data without explicit instructions, potentially uncovering relationships that may not be immediately obvious to humans. Following this logic, this ability to find latent structures could be advantageous in capturing complex structural information in graph data and lead to similar breakthought than discovering new particle physics phenomena using clustering and dimensionality reduction on collider data or identifying subtypes of cancer with clustering and PCA in genomic studies.
Using the \emph{ClinTox Molecular} dataset to assess molecular toxicity, we explored how key graph properties, such as the \textit{average degree} and \textit{spectral radius}, are utilized by our GIN architecture. The average degree, closely linked to atomic valency, reflects a molecule's potential for interactions. The \textit{spectral radius} offers a complementary hypothesis, suggesting that the overall structural stability of a molecule, independent of specific atomic features, may also be a key factor in toxicity prediction.

\section{Future work}
Our methodology has several limitations. While we addressed dataset issues such as leakage and isomorphic graphs, a key challenge remains the lack of guarantees that GNNs find globally optimal solutions, despite their theoretical capacity as universal function approximators \citep{HORNIK1989359}.
As we observed, early layers in the network contain predictive information for some global graph features. It is however unclear why this is the case. They might be somehow predictive for the features we are looking for, perhaps in the specific combination with the specific distribution of graphs we used int he experiments. 
% This is particularly evident in fMRI data, where multiple layers of complexity—from MRI limitations and BOLD signal characteristics to Pearson correlation for functional connectivity—introduce noise and inaccuracies. 
Investigating additional graph properties like girth or complex motifs could be beneficial. Preliminary work on alternative architectures (e.g., GATv2, GraphSAGE, ChebNet, Set2Set, HO-Conv, DiffPool) has begun but is not yet complete. An extensive exploration of 1-WL, 2-WL and 3-WL GNN equivalent could bolster the paper's contributions by showing clear restrictions and capabilities of these models. As a future encouraging work, studying the supervision signal comparing unsupervised, self-supervised and supervised models would be very insightful. Interesting application research could be done on Knowledge Graphs. Another path could be fMRI connectomes, which are also graphs. With fMRI data, some functional connectivity patterns could be associated with cognitive states under the lens of GNNs probing.

\section{Conclusion}

We demonstrate the relevance of using a probing classifier as a model-agnostic explainability method for graph neural networks. We manifest both the expressivity of different GNN architectures and their capacity to solve a graph classification problem through effective feature extraction. They render it linearly separable in the space of their embeddings through the computation of graph properties. We validate domain knowledge with the Clintox Molecular dataset. There is a manifest emergence of molecular qualities like toxicity with regard to their structural properties like \textit{node degree} (atom valency) and \textit{spectral radius} (the molecule's stability). To explain a macro attribute, there are instances where structural properties may offer more insight than the mere aggregation of element properties. This provide encouraging results for studying the possibility of formulating hypotheses on the emergent dependence of complex systems attributes to basic and more higher level structural properties. We could explore how the macroscopic behavior of complex systems emerges from the intricate interplay of their microscopic components, raising hypothesis on graph similarly than what is now done on language with LLMs or on vision with computer vision.

% I'd like to conclude this master's thesis with a citation of one of my favourite researcher~\citep{freeman2008nonlinear} “On the one hand, philosophers do not understand nonlinear brain dynamics well enough to adjudicate conflicting claims of neuroscientists working on opposing sides of the cleavage: linear-passive versus nonlinear-active perception. On the other hand, most neuroscientists do not understand the philosophical foundations of brain theory well enough to focus their experimental questions in terms of the self-organisation of brains in behaviour, or even to know which side of the cleavage they are on, or that it exists”. Following this statement, it appears natural to me that discoveries and interesting science can emerge from filling the gaps between trans-disciplinary research.

\newpage

\section*{Acknowledgment}
This work is based on the MSc. AI thesis by Tom Pelletreau-Duris, a large part of the results were also presented in that work. 

Michael Cochez is partially funded by the Elsevier Discovery Lab, partially funded by the Graph-Massivizer project, funded by the Horizon Europe programme of the European Union (grant 101093202), and supported by a gift from Accenture LLP. His work on this publication is in part based upon work from COST Action CA23147 GOBLIN - Global Network on Large-Scale, Cross-domain and Multilingual Open Knowledge Graphs, supported by COST (European Cooperation in Science and Technology, https://www.cost.eu).

We would also want to thank the anonymous reviewers for their comments and suggestions that helped us improve the manuscript.

\subsubsection*{Author Contributions}

Pelletreau-Duris Tom: Conceptualization (lead); methodology (lead); software (lead); formal analysis (lead); writing – original draft (lead); writing – review and editing (equal).  
Ruud van Bakel: writing – review and editing (equal); 
Michael Cochez: writing – review and editing (equal); Supervision; Funding acquisition

% \acks{Acknowledgements go here.}
%RvB and MC are in part funded by the Graph-Massivizer project, the Horizon Europe research and innovation program of the European Union grant management number 101093202 \url{https://graph-massivizer.eu/}

\bibliography{refs.bib}
\newpage

\appendix

\section{\texorpdfstring{R\textsuperscript{2}}{R2} Score} 
\label{R2score}
We are using $R\textsuperscript{2}$ as the main metrics. The $R\textsuperscript{2}$ score (coefficient of determination) measures the proportion of variance in the dependent variable that is predictable from the independent variable(s). For a probing classifier, it would indicate how well the probe’s predictions match the actual properties being probed. More formally, $R\textsuperscript{2}$ is defined as:

$$
R^2 = 1 - \frac{\sum_{i} \left( z^{(i)} - \hat{z}^{(i)} \right)^2}{\sum_{i} \left( z^{(i)} - \bar{z} \right)^2}
$$

Where:
$z^{(i)}$ is the ground truth value of the property $\hat{z}$ for the $i$-th data point in the probing dataset $\mathcal{D}_{P}$.
$\hat{z}^{(i)}$ is the predicted value of $\hat{z}$ produced by the probing classifier $g$.
$\bar{z}$ is the mean of the ground truth values $z^{(i)}$ over the dataset.
The numerator represents the residual sum of squares (how far off the predictions are), and the denominator represents the total sum of squares (the variance in the ground truth values).

An $R\textsuperscript{2}$ value ranges from 0 to 1, where: $R\textsuperscript{2} = 1$ means the probing classifier perfectly explains the variance in the target property (i.e., the learned representations fully capture the property). $R\textsuperscript{2} = 0$ means the probing classifier does no better than predicting the mean $\bar{z}$, implying the representations do not capture any useful information about the property. Good R2 score should indicate how the model achieves its behavior on the original task \cite{hupkes2018visualisation}.

A good $R\textsuperscript{2}$ score gives a sense of how well the features at each layer can be separated linearly to predict the target labels. The second reason is that a more complex probe “bears the risk that the classifier infers features that are not actually used by the network”~\citep{hupkes2018visualisation}. Of course, other non linear probes have been explored in the literature~\citep{belinkov2021probing}. If a few studies observed better performance with more complex probes, the logic remained the same $\operatorname{Perf}\left(g, f_1, \mathcal{D}_O, mathcal{D}_P\right)>\operatorname{Perf}\left(g, f_2, \mathcal{D}_O, \mathcal{D}_P\right)$, of two representations $f_1(x)$ and $f_2(x)$, holds across different probes $g$. The important criteria is to compare the results obtained by the same measurement system. In general, if we can predict one property on one embedding for a given classification problem, then it means this properly is useful for the problem resolution.

From an information-theoretic perspective, training the probing classifier $g$ can be viewed as estimating the mutual information between the learned representations $f_{l}(x)$ and the property $z$. This mutual information is denoted as $\mathrm{I}(\mathbf{z}; \mathbf{h})$, where $\mathbf{z}$ refers to the property and $\mathbf{h}$ represents the intermediate representations~\citep{belinkov2021probing}.
% END

\newpage
%\subsection*{Appendix}

\section{Local and global graph properties}

\label{sec:appendix:graph_properties}

\begin{table}[H]
\centering
\small
\begin{tabular}{|c|>{\centering\arraybackslash}p{2cm}|>{\centering\arraybackslash}p{6cm}|>{\centering\arraybackslash}p{5cm}|}
\hline
\multirow{6}{*}[0pt]{\rotatebox[origin=c]{90}{\textbf{Local}}} & \textbf{Property} & \textbf{Visual Pattern \& Definition} & \textbf{Computational Criteria} \\
\cline{2-4}
& Degree & 
How many links a node has which is the simplest form of centrality  & 
Count edges per node \\
\cline{2-4}
& Local clustering Coefficient & 
Are the neighbours of a node also connected together ? & 
Count triangles of neighbours / total possible triangles of neighbours \\
\cline{2-4}
& Betweenness Centrality & 
How much of a bridge between clusters is a node. Removing that node would break many shortest paths. Importance in information flow & 
Number of shortest paths through node \\
\cline{2-4}
& Closeness Centrality & 
Being in the middle of the network, the barycenter of the graph.  & 
The average length of the geodesic distances to all the other nodes (inverse sum of shortest paths) \\
\cline{2-4}
& Eigenvector Centrality & 
Being connected to well connected nodes without necessarily having a large number of neighbours itself; influence based on connections & 
Recursive definition based on neighbours \\
\cline{2-4}
& PageRank & 
Nodes with important connections; web-inspired importance & 
Similar to Eigenvector but with random walk and teleportation \\
\hline
\end{tabular}
\caption{Local Network Properties with definition and computational criteria}
\label{tab:network_properties_combined_1}
\end{table}

\newpage
%\clearpage

\begin{table}[H]
\centering
\small
\renewcommand{\arraystretch}{1.3} % Adjusts row height
\begin{tabular}{|c|p{4cm}|p{5cm}|p{5cm}|}
\hline
\multirow{18}{*}[0pt]{\rotatebox[origin=c]{90}{\textbf{Global}}} & \textbf{Property} & \textbf{Visual Pattern \& Definition} & \textbf{Computational Criteria} \\
\cline{2-4}
& Number of Nodes & Graph size; total nodes in the network & Count vertices \\
\cline{2-4}
& Number of Edges & Graph density; total connections in the network & Count connections \\
\cline{2-4}
& Density & Overall graph connectivity; how densely connected & Ratio of actual to possible edges \\
\cline{2-4}
& Average Path Length & On average, how close are nodes to each other? Typical distance between node pairs & Average number of steps along the shortest paths for all possible pairs of nodes \\
\cline{2-4}
& Diameter & Graph span; longest of all shortest paths & Maximum shortest path \\
\cline{2-4}
& Radius & Graph core; minimum distance from central to farthest node & Minimum eccentricity \\
\cline{2-4}
& Transitivity & Triangle density; probability of connected node triplets & Ratio of triangles to triads \\
\cline{2-4}
& Assortativity & Node degree correlations; tendency of similar nodes to connect & Pearson correlation of degrees \\
\cline{2-4}
& Number of Cliques & Dense subgraphs; count of maximal fully connected subgraphs & Number of maximal complete subgraphs \\
\cline{2-4}
& Number of Triangles & Local density; fully connected 3-node subgraphs & Count 3-node cliques \\
\cline{2-4}
& Number of Squares & 4-node patterns; cycles in the graph & Count 4-node cycles \\
\cline{2-4}
& Largest Component Size & Main connected structure; size of biggest connected part & Largest set of connected nodes \\
\cline{2-4}
& Average Degree & Overall connectivity; average connections per node & Mean of all node degrees \\
\cline{2-4}
& Spectral Radius & Dominant graph structure; overall connectivity measure & Largest eigenvalue of adjacency matrix \\
\cline{2-4}
& Algebraic Connectivity & Graph cohesion; measure of how well-connected the graph is & Second smallest eigenvalue of Laplacian \\
\cline{2-4}
& Graph Energy & The eigenvalues capture deviations from regularity in the network. Complete graphs or highly connected networks tend to have higher energies due to the larger magnitude of their eigenvalues. Graph energy can help assess robustness, synchronizability. & Sum of absolute Laplacian eigenvalues \\
\cline{2-4}
& Small World Coefficient & Balance of clustering and paths; small-world characteristics & Comparison to random graph \\
\cline{2-4}
& Small World Index & Refined small-world measure; comparison to random and lattice graphs & Comparison to random and lattice graphs \\
\cline{2-4}
& Betweenness Centralization & Central node dominance; degree of central bridging node & Variation in betweenness centrality across nodes \\
\cline{2-4}
& PageRank Centralization & Influence concentration; degree of dominant influential nodes & Variation in PageRank values across nodes \\
\hline
\end{tabular}
\caption{Global Network Properties with definition and computational criteria}
\label{tab:network_properties_combined_2}
\end{table}

We are using the Small-World Index, $SWI = \left( \frac{L - L_l}{L_r - L_l} \right) \times \left( \frac{C - C_r}{C_l - C_r} \right)$ in our experiment because it provides a more balanced and robust measure of small-world properties. Unlike the Small-World Quotient: $Q = \frac{C / C_r}{L / L_r}$, which can be sensitive to network size and degree, $SWI$ normalizes both the clustering coefficient and average path length with respect to both random and lattice reference graphs. This dual normalization approach ensures that $SWI$ is less prone to false positives or negatives, making it a more reliable metric for our analysis \citep{neal2017how}. 

\section{Literature review on related Work}
\label{relatedwork}
Existing post-hoc GNN explanations methods can be classified into two main categories: \textit{instance-level} and \textit{model-level} methods~\citep{barredo_arrieta2020explainable}. See~\citep{agarwal2023evaluating,dai2022comprehensive} for nice reviews on the subject. 
%
% maybe to put away who knows ? 
In the realm of instance based methods, \textit{gradient-based} methods use the gradients of the output with respect to the input or intermediate features to measure the importance of each component of the graph. \textit{Decomposition-based} methods try to decompose the input graph into smaller subgraphs or paths that can account for the output. \textit{Surrogate-based} methods use a simpler, more interpretable model to approximate the behavior of the original GNN and provide explanations based on the surrogate model. And finally \textit{Perturbation-based} methods which perturb the input graph by removing or adding nodes, edges, or features, and observe the changes in the output to identify the influential components. The most mainstream technique, GNNExplainer~\citep{ying2019gnnexplainer} achieves explanation by removing redundant edges from an input graph instance, maximizing the mutual information between the distribution of subgraphs and the GNN’s prediction. It is able to provide an explanation both in terms of a subgraph of the input instance to explain, and a feature mask indicating the subset of input node features which is most responsible for the GNN’s prediction.

For \textit{model-based} techniques, few methods come to mind~\citep{saha2022model-centric,azzolin2023globalexplainabilitygnnslogic,vu2020pgm-explainer,wang2023reinforced,xuanyuan2023globalconceptbasedinterpretabilitygraph,yuan2020xgnn,zhang2021relex}. The most mainstream method seems to be XGNN~\citep{yuan2020xgnn}. The authors of XGNN investigate the possible input characteristics used by a GNN for graph classification. But they formulate the problem as a reinforcement learning problem and generate graph patterns iteratively. Such an iterative approach is often intractable for large graphs. Moreover, it does not allow for both node classification and graph classification explanations, nor does it allow for an investigation of the learning process through the different layers of the GNN. In general, none of the techniques allow for an interpretation of the hidden representations states with graph properties.

\newpage

\newpage

\section{Grid House artificial dataset}

\subsection{Grid House figures} \phantom{.}
% \todo{Add a legend to this figure. It is now not clear what colors and node size indicate.}

\begin{figure}[ht]
    \centering
    \begin{minipage}[b]{0.3\textwidth}
        \centering
        \includegraphics[width=\textwidth]{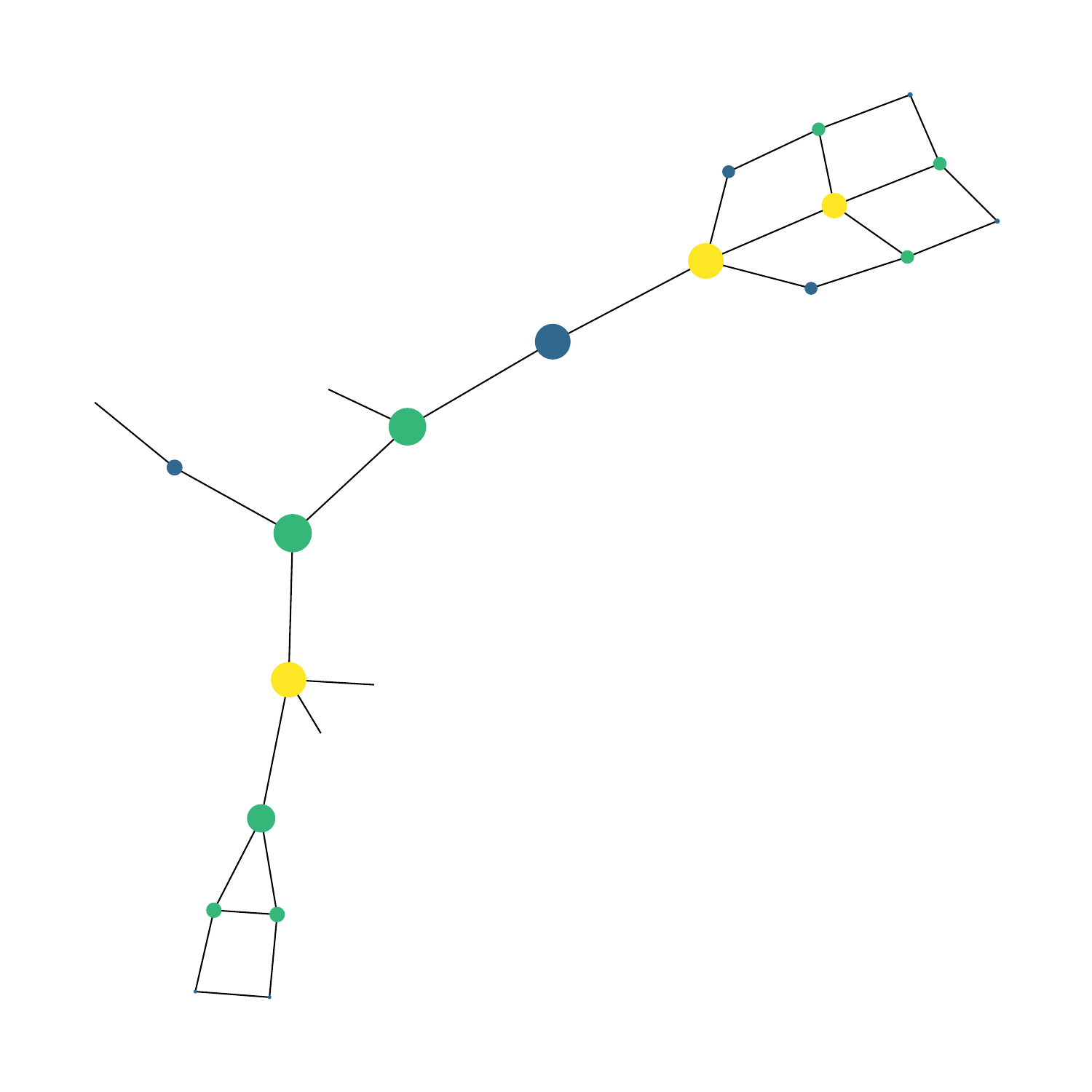}
    \end{minipage}
    \hfill
    \begin{minipage}[b]{0.3\textwidth}
        \centering
        \includegraphics[width=\textwidth]{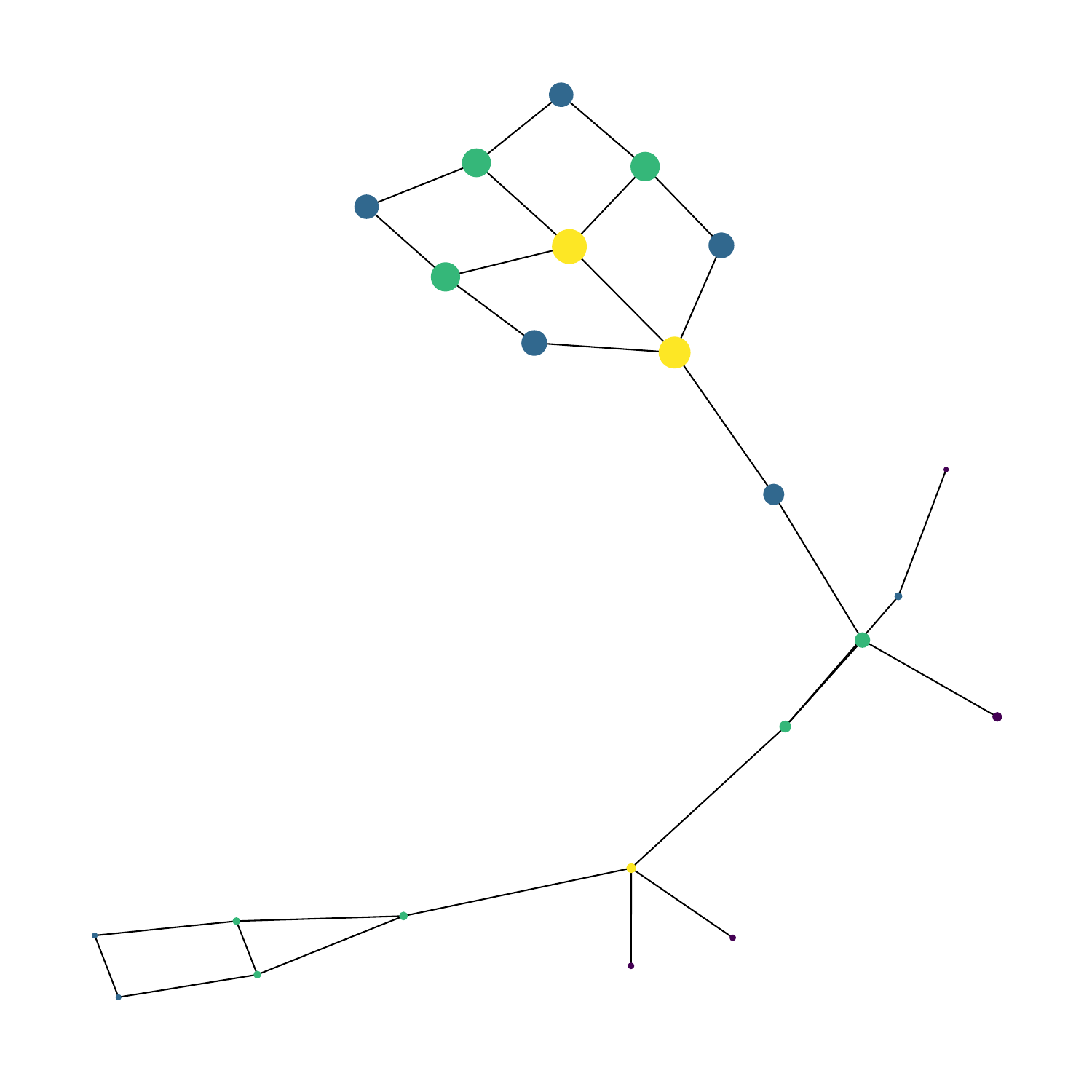}
    \end{minipage}
    \hfill
    \begin{minipage}[b]{0.3\textwidth}
        \centering
        \includegraphics[width=\textwidth]{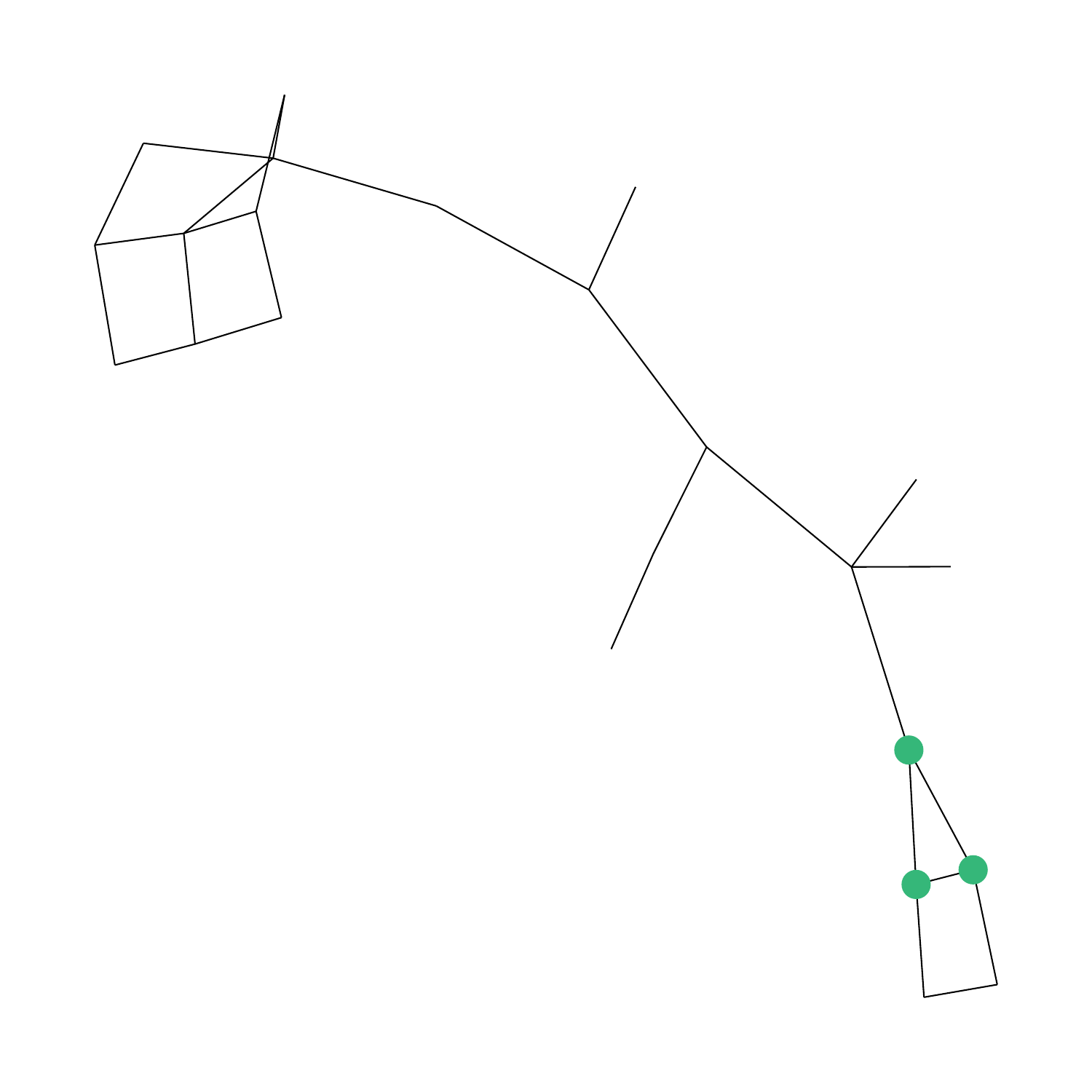}
    \end{minipage}
    \caption{Comparison of different centrality measures for the first graph in our Grid House dataset: (a) betweenness centrality, (b) eigenvector (PageRank) centrality, and (c) local clustering coefficients.}
    \label{fig:centrality_measures}
\end{figure}

\clearpage

Experiments \footnote{All the experiments have been done on the dutch supercomputer SNELLIUS (SURF) using GPU and CPU.  }

\subsection{Grid House models} \phantom{.} 

\begin{table}[htbp]
    \caption{Range of Hyper-parameters and Final Specification for the Grid-House Dataset}
    \label{table:brainib_hyperparameters}
    \centering
    \begin{tabular}{ccc}
        \hline 
        Hyper-parameter & Range Examined & Final Specification \\
        \hline 
        Graph Encoder & & \\
        \hline 
        \#GNN Layers & \{$[2,3,4,5]$\} & 4 (GCN), 2 (GIN), 3 (GAT) \\
        \#MLP Layers & \{$[2,3,4]$\} & 3 (GCN), 2 (GIN), 2 (GAT) \\
        Hidden Dimensions & \{$[10, 15, 30, 45, 60, 64, 128, 256]$\} & 60 (GCN), 30 (GIN), 128 (GAT) \\
        Attention Heads (GAT) & \{$[4, 8, 16]$\} & 8 heads, 32 dimensions per head \\
        \hline 
        Learning Rate & \{$[1 \mathrm{e}-2,1 \mathrm{e}-3,1 \mathrm{e}-4]$\} & $1 \mathrm{e}-3$ \\
        Batch Size & \{$[32,64,128, 256]$\} & 64 \\
        Weight Decay (when added) & \{$[1 \mathrm{e}-4,1 \mathrm{e}-2]$\} & $1\mathrm{e}-4$ (GCN), $1\mathrm{e}-2$ (GIN)  \\
        Batch Normalization & \{$with, without$\} & $without$ \\
        Dropout (when added) & \{[0.15, 0.5]\} & 0.2 \\
        Pooling Method & \{$mean, sum, max$\} & $max$ (GCN), $mean$ (GIN), $max$ (GAT) \\
    \end{tabular}
\end{table}
% \vspace{-50pt}

\begin{table}[htbp]
    \caption{Performance of Different Models with Regularization on the Artificial Dataset (80\%-20\% Random Split). The highest performance is highlighted with boldface. All performances are reported under their best settings and rounded to 2 decimal places.}
    \label{table:gridhouseperformance}
    \centering
    \begin{tabular}{ccc}
        \hline
        \textbf{Method}  & \textbf{Test Accuracy} \\
        \hline
        GCN (control) & $0.90$ \\
        GCN ($L_2$) & $0.97$ \\
        GCN (dropout) & $0.93$ \\
        GIN (control) & $\mathbf{1.00}$ \\
        GIN ($L_2$) & $0.99$ \\
        GIN (dropout) & $1.00$ \\
        GAT & $0.97$ \\
        \hline
    \end{tabular}
\end{table}

As expected the RGCN outperform the GCN on this node classification task.

\begin{figure}[htbp]
    \centering
    \includegraphics[width=0.78\linewidth]{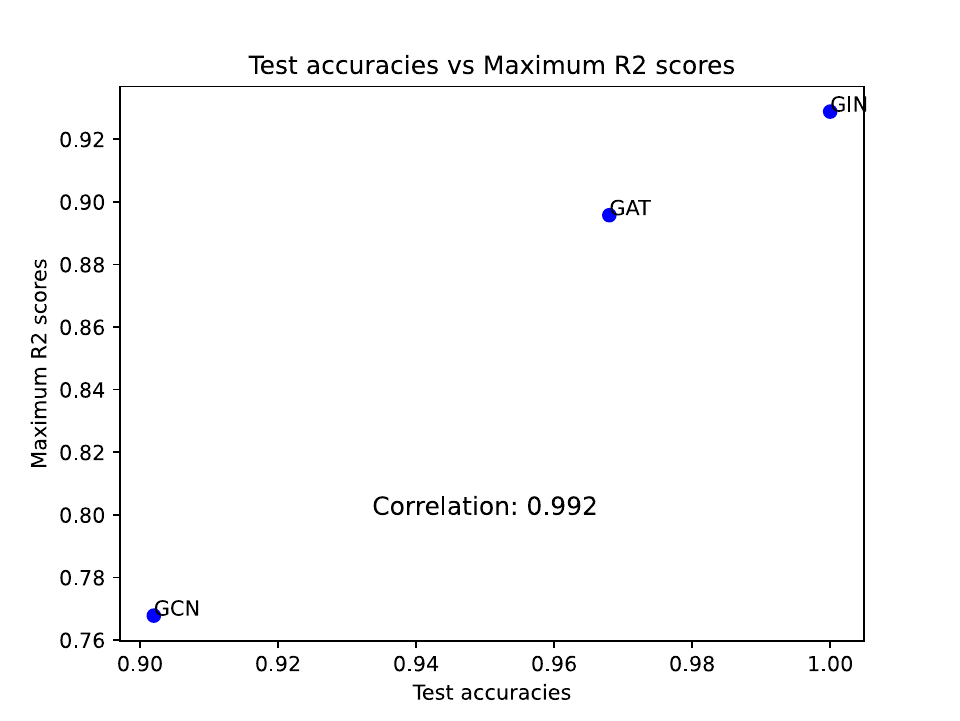}
    \caption{Plot of the correlation between the different model test accuracies and their maximum R2 score (Grid House)}
    \label{fig:correlation}
\end{figure}

% \subsubsection{AIFB models} \phantom{.} % Adds an invisible character, creating vertical space
% In this section we only worked on node classification and node probing which helped us develop a method for the node probing technique which we further applied to the Grid-House dataset and the fMRI dataset. 
% \begin{table}[htbp]
%     \caption{Performance of Different Models on the artificial dataset with a 80\%-20\% Random Split. The highest performance is highlighted with boldface. All the performance of methods are reported under their best settings.}
%     \label{table:cross_validation_performance}
%     \centering
%     \begin{tabular}{ccc}
%         \hline
%         \textbf{Model}  & \textbf{AIFB (Acc)} \\
%         \hline
%         GCN & $0.89$ \\
%         RGCN & $0.94$ \\
%         \hline
%     \end{tabular}
% \end{table}

\clearpage

\subsection{Grid House RESULTS}

\subsubsection{Graph properties probing results} \phantom{.} % Adds an invisible character, creating vertical space

\begin{table}[ht]
\centering
\caption{Linear Probing $R^2$ Performance Across models for Selected Graph Properties (GridHouse Dataset). Best Scores in Bold; Non-convergence indicated by \textemdash } % Linear}
\resizebox{\textwidth}{!}{
    \begin{tabular}{lccccccccc}
    \toprule
    \textbf{Model} & \#nodes & \#edges & density & avg path len & \#cliques & \#triangles & \#squares & \#Largest Component \\
    \midrule
    \textbf{GCN (control)} \\
    \midrule
    x\_global & 0.36 & \textemdash & 0.66 & 0.33 & 0.02 & 0.31 & \textbf{0.77} & 0.36 \\
    x5 & 0.33 & 0.22 & 0.64 & 0.29 & 0.27 & 0.39 & \textbf{0.77} & 0.33 \\
    x6 & 0.19 & 0.08 & 0.56 & \textemdash & 0.07 & 0.06 & \textbf{0.74} & 0.19 \\
    x7 & \textemdash & \textemdash & 0.45 & 0.13 & \textemdash & 0.03 & \textbf{0.72} & \textemdash \\
    \midrule
    \textbf{GCN ($L_2$)} \\
    \midrule
    x\_global & 0.36 & 0.09 & 0.67 & 0.35 & 0.20 & 0.68 & \textbf{0.86} & 0.36 \\
    x5 & 0.31 & 0.32 & 0.66 & 0.32 & 0.32 & 0.80 & \textbf{0.86} & 0.31 \\
    x6 & 0.04 & \textemdash & 0.41 & 0.15 & 0.03 & 0.23 & \textbf{0.83} & 0.04 \\
    x7 & \textemdash & \textemdash & 0.29 & 0.27 & \textemdash & 0.09 & \textbf{0.81} & \textemdash \\
    \midrule
    \textbf{GCN (dropout)} \\
    \midrule
    x\_global & 0.21 & 0.07 & 0.67 & 0.33 & 0.07 & 0.63 & \textbf{0.72} & 0.22 \\
    x5 & \textemdash & \textemdash & 0.59 & 0.26 & \textemdash & 0.66 & \textbf{0.74} & \textemdash \\
    x6 & \textemdash & \textemdash & 0.42 & 0.21 & \textemdash & 0.49 & \textbf{0.65} & \textemdash \\
    x7 & \textemdash & \textemdash & 0.35 & 0.10 & \textemdash & 0.26 & \textbf{0.51} & \textemdash \\
    \midrule
    \textbf{GIN (control)} \\
    \midrule
    x\_global & 0.12 & 0.07 & 0.50 & 0.32 & 0.07 & 0.22 & \textbf{0.87} & 0.12 \\
    x5 & \textemdash & \textemdash & 0.72 & 0.30 & \textemdash & 0.89 & \textbf{0.93} & \textemdash \\
    x6 & \textemdash & \textemdash & \textemdash & 0.02 & \textemdash & 0.11 & \textbf{0.88} & \textemdash \\
    \midrule
    \textbf{GIN ($L_2$)} \\
    \midrule
    x\_global & \textemdash & \textemdash & 0.49 & 0.30 & \textemdash & 0.18 & \textbf{0.85} & \textemdash \\
    x5 & \textemdash & \textemdash & 0.51 & 0.15 & \textemdash & 0.52 & \textbf{0.89} & \textemdash \\
    x6 & \textemdash & \textemdash & 0.40 & 0.12 & \textemdash & 0.10 & \textbf{0.80} & \textemdash \\
    \midrule
    \textbf{GIN (dropout)} \\
    \midrule
    x\_global & \textemdash & \textemdash & 0.53 & 0.36 & \textemdash & 0.25 & \textbf{0.87} & \textemdash \\
    x5 & \textemdash & \textemdash & 0.71 & 0.33 & \textemdash & 0.85 & \textbf{0.93} & \textemdash \\
    x6 & \textemdash & \textemdash & \textemdash & 0.21 & \textemdash & 0.34 & \textbf{0.91} & \textemdash \\
    \midrule
    \textbf{GAT} \\
    \midrule
    x\_global & 0.54 & 0.59 & \textemdash & 0.49 & 0.61 & \textbf{0.89} & 0.87  & 0.54 \\
    x5 & \textemdash & \textemdash & 0.33 & 0.27 & \textemdash & 0.17 & \textbf{0.64} & \textemdash \\
    x6 & \textemdash & \textemdash & 0.25 & 0.17  & \textemdash & 0.17 & \textbf{0.63} & \textemdash \\
    \bottomrule
    \end{tabular}
    }
\label{tab:gridhousegraphprobing}
\end{table}

\clearpage
\newpage

\subsubsection{Graph properties probing plots} \phantom{.} % Adds an invisible character, creating vertical space

\textbf{GCN}

\begin{figure}[h]
    \centering
    \includegraphics[width=0.9\linewidth]{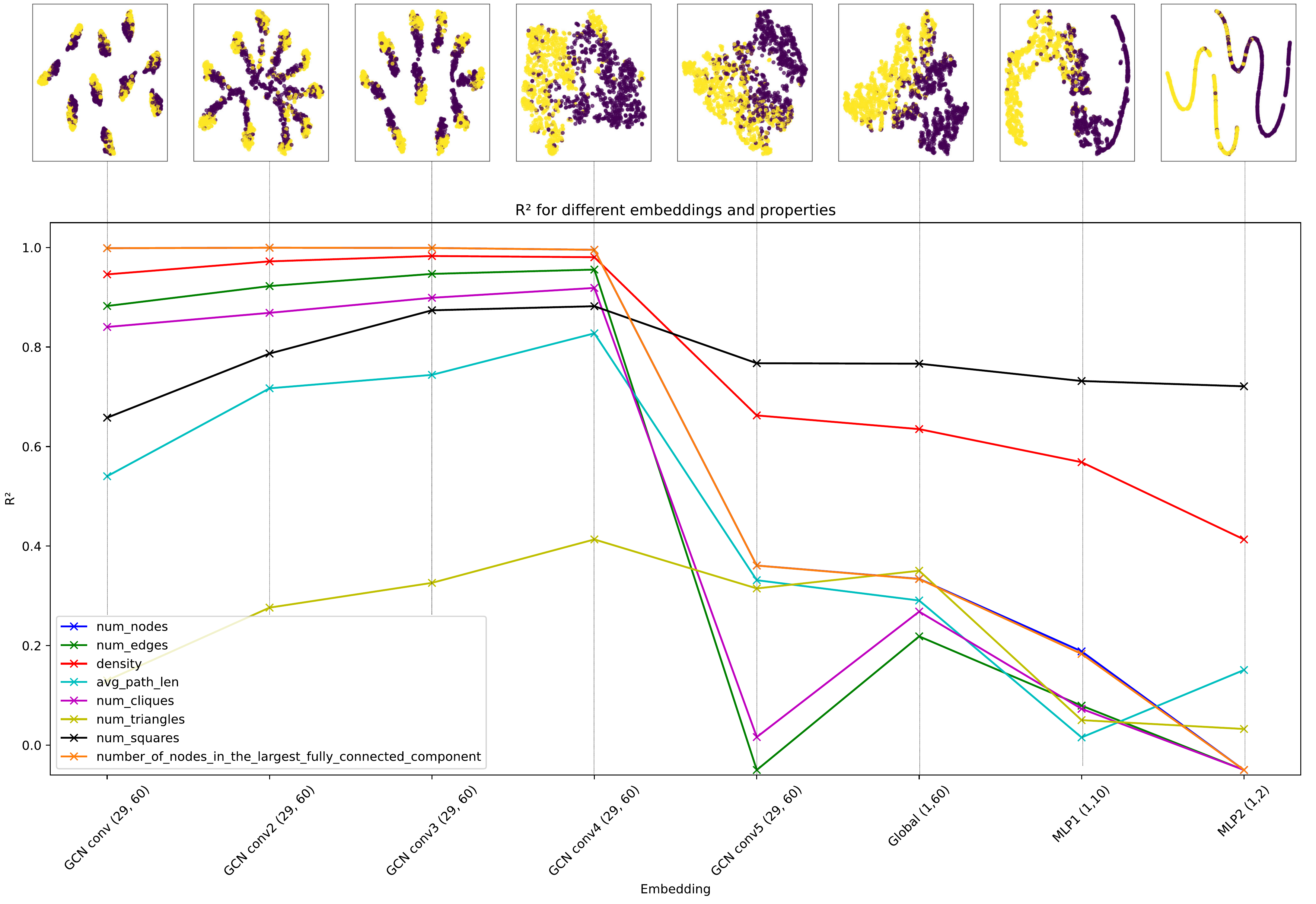}
    \caption{T-SNE visualization across different
layers of our GCN architecture aligned with the probing $R^2$ scores plots (Grid House)}
    \label{fig:plotGCNTSNEcontrol}
\end{figure}

\begin{figure}[h]
    \centering
    \includegraphics[width=0.9\linewidth]{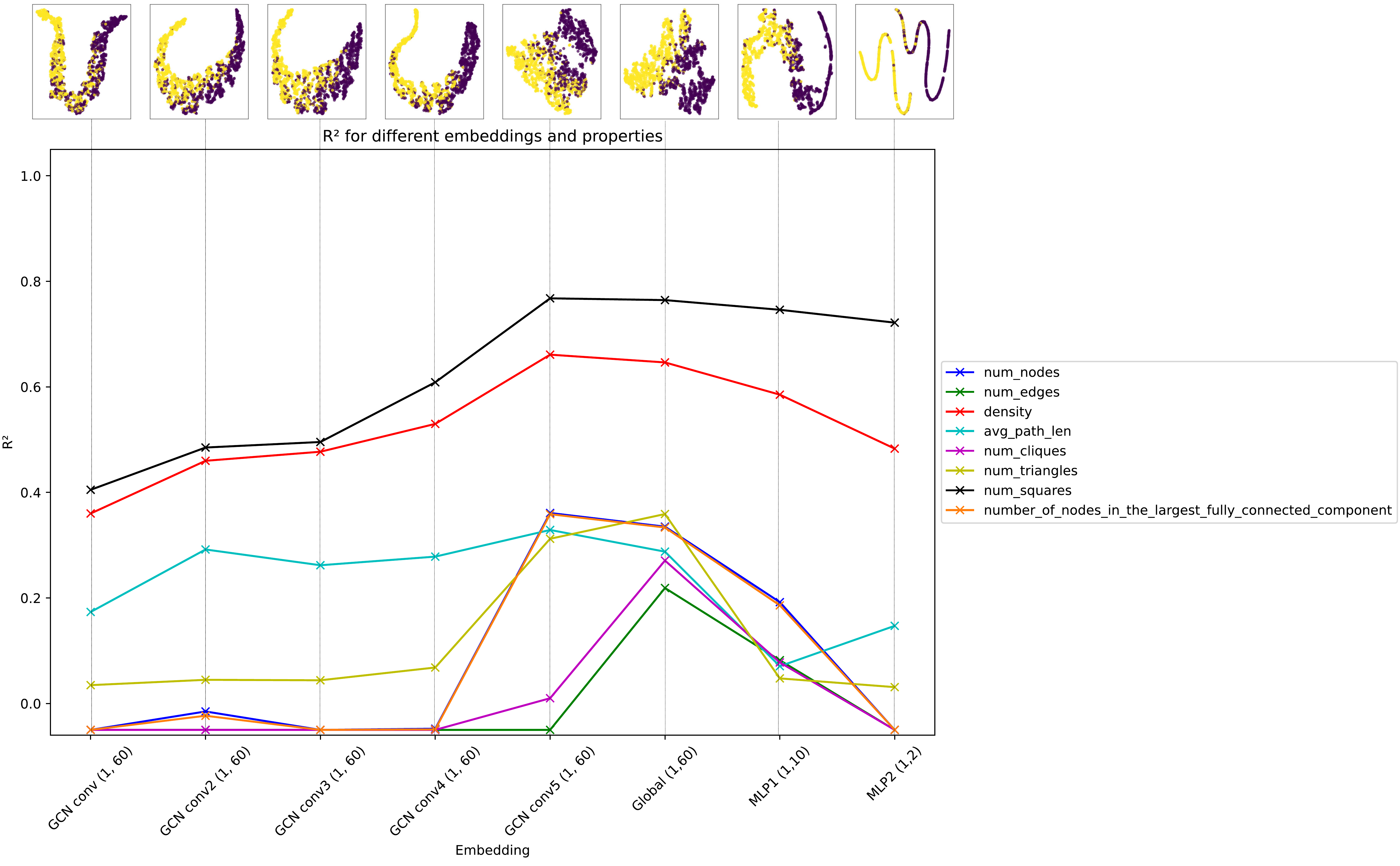}
    \caption{T-SNE visualization across different
layers of our GCN architecture aligned with the probing $R^2$ scores plots with mean-pooled node embeddings (Grid House)}
    \label{fig:plotGCNTSNEcontrol_pooled}
\end{figure}

% \begin{figure}
%     \centering
%     \includegraphics[width=1\linewidth]{figures/BA_2grid_house_no_isomorphic_GCN_tsne.pdf}
%     \caption{T-SNE visualisation across different
% layers of our GCN architecture (Grid House)}
%     \label{fig:plotGCNTSNEcontrol}
% \end{figure}

% \begin{figure}
%     \centering
%     \includegraphics[width=1\linewidth]{BA_2grid_house_no_isomorphic_GCN_R2_all_layers.pdf}
%     \caption{Plot of the GCN (control) R2 results across different
% layers probing for graph properties (Grid House)}
%     \label{fig:plotGCNcontrol_full}
% \end{figure}

\begin{figure}[h]
    \centering
    \includegraphics[width=1\linewidth]{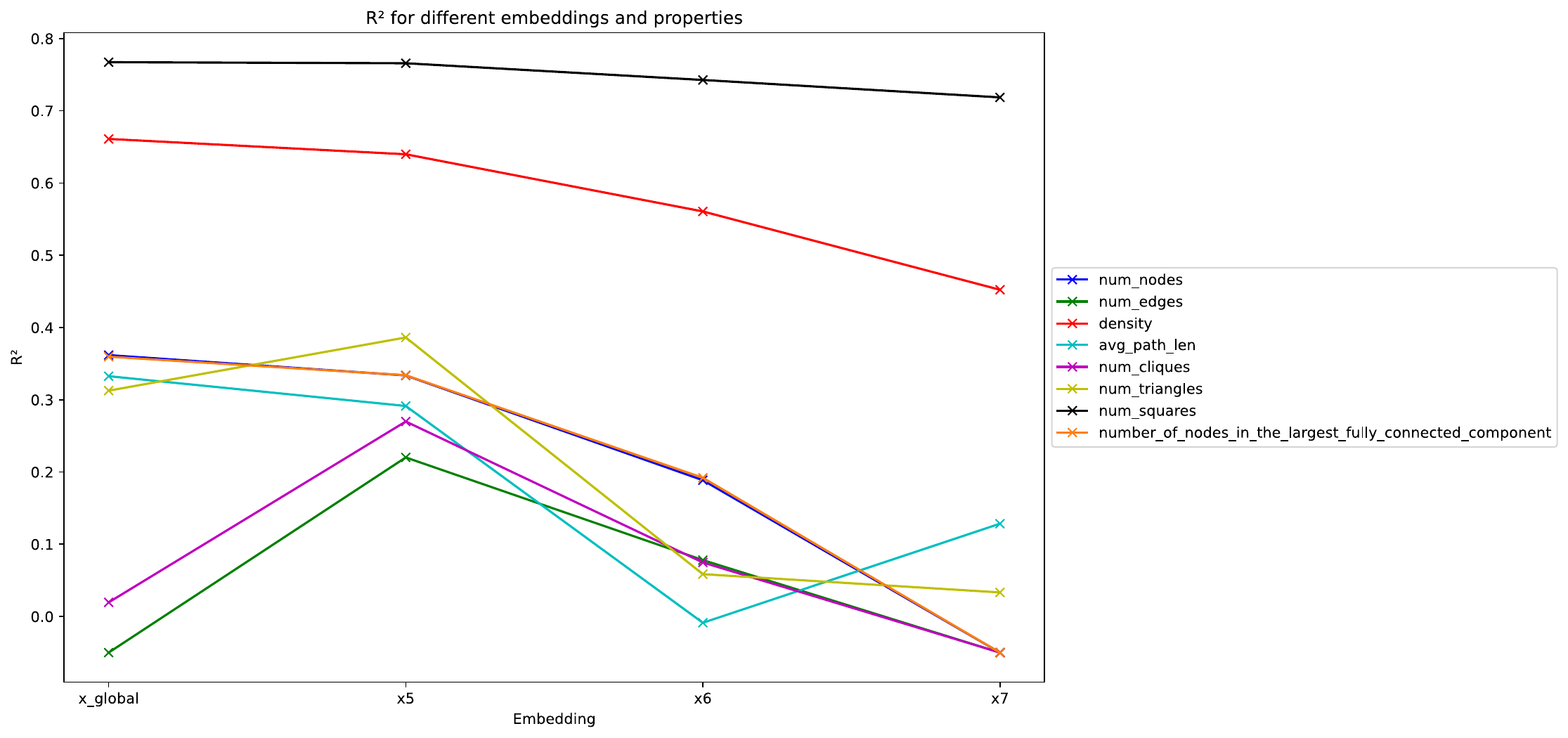}
    \caption{Plot of the GCN (control) $R^2$ results across different
layers probing for graph properties with post pooling layers only, allowing clearer visualization and higher order property interpretation (Grid House)}
    \label{fig:plotGCNcontrol}
\end{figure}

\begin{figure}[h]
    \centering
    \includegraphics[width=1\linewidth]{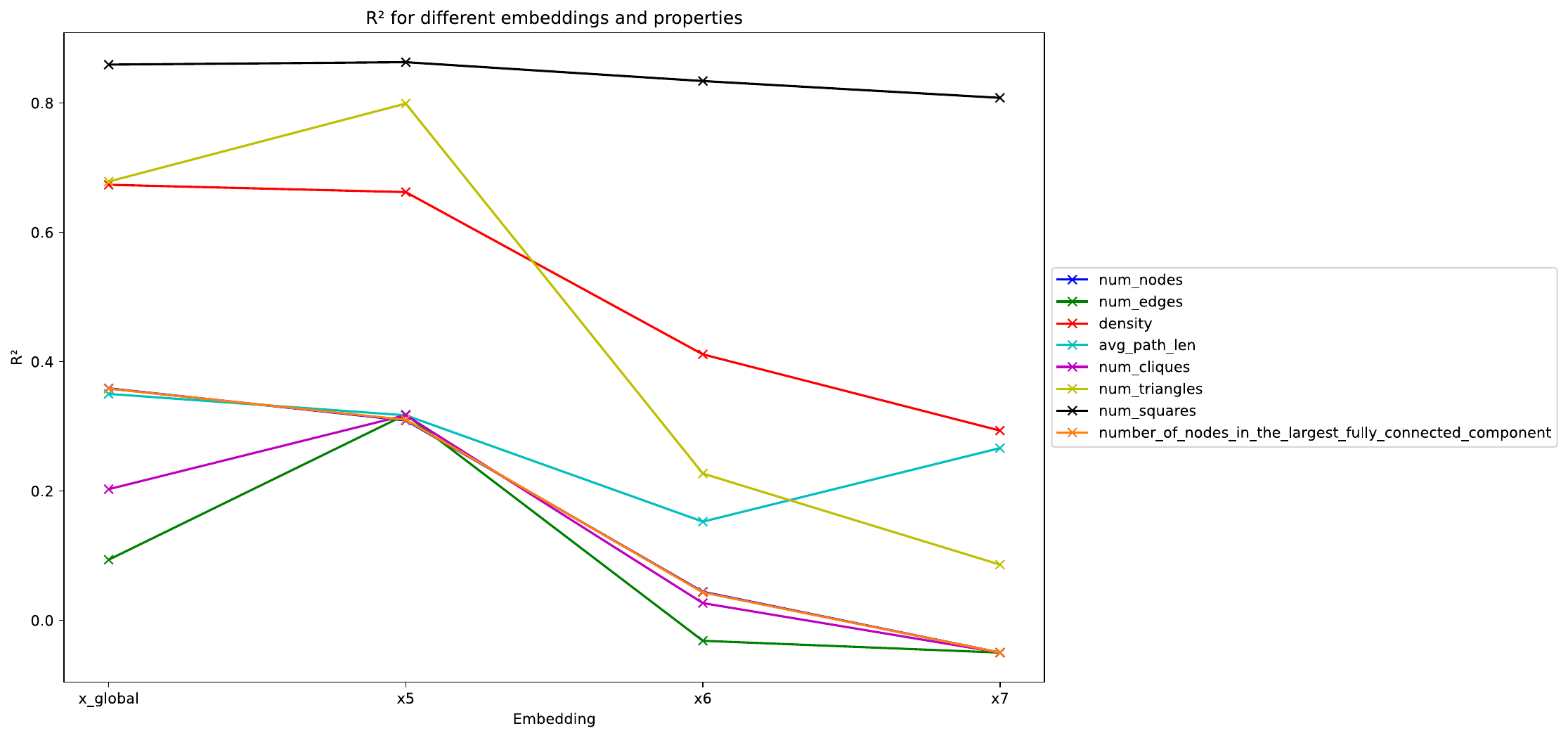}
    \caption{Plot of the GCN ($L_2$) $R^2$ results across different
layers probing for graph properties with post pooling layers only, allowing clearer visualization and higher order property interpretation (Grid House)}
    \label{fig:plotGCNL2}
\end{figure}

\begin{figure}[h]
    \centering
    \includegraphics[width=1\linewidth]{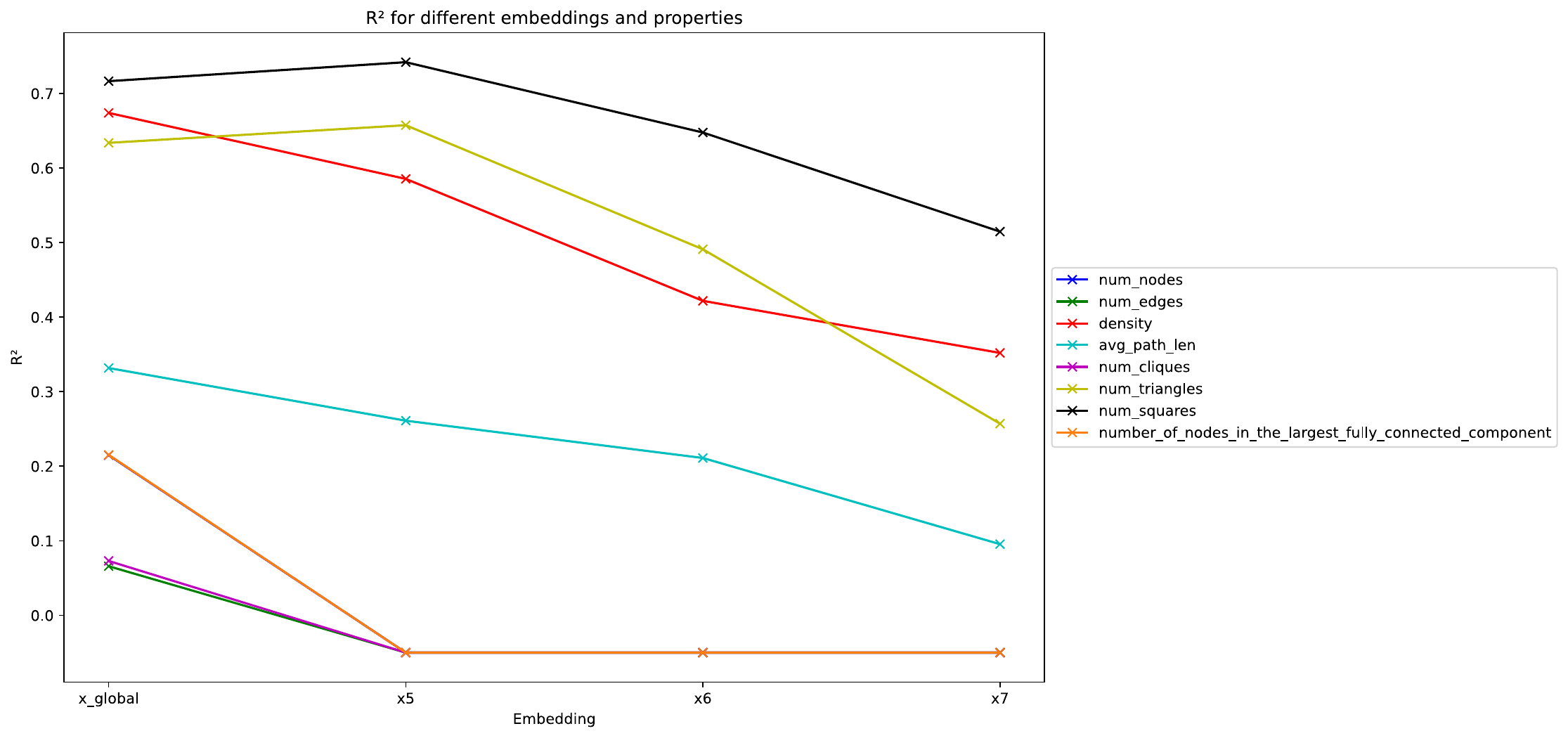}
    \caption{Plot of the GCN (dropout) $R^2$ results across different
layers probing for graph properties with post pooling layers only, allowing clearer visualization and higher order property interpretation (Grid House)}
    \label{fig:plotGCNLdropout}
\end{figure}

\newpage
\clearpage

\textbf{GIN}

\begin{figure}[h]
    \centering
    \includegraphics[width=0.9\linewidth]{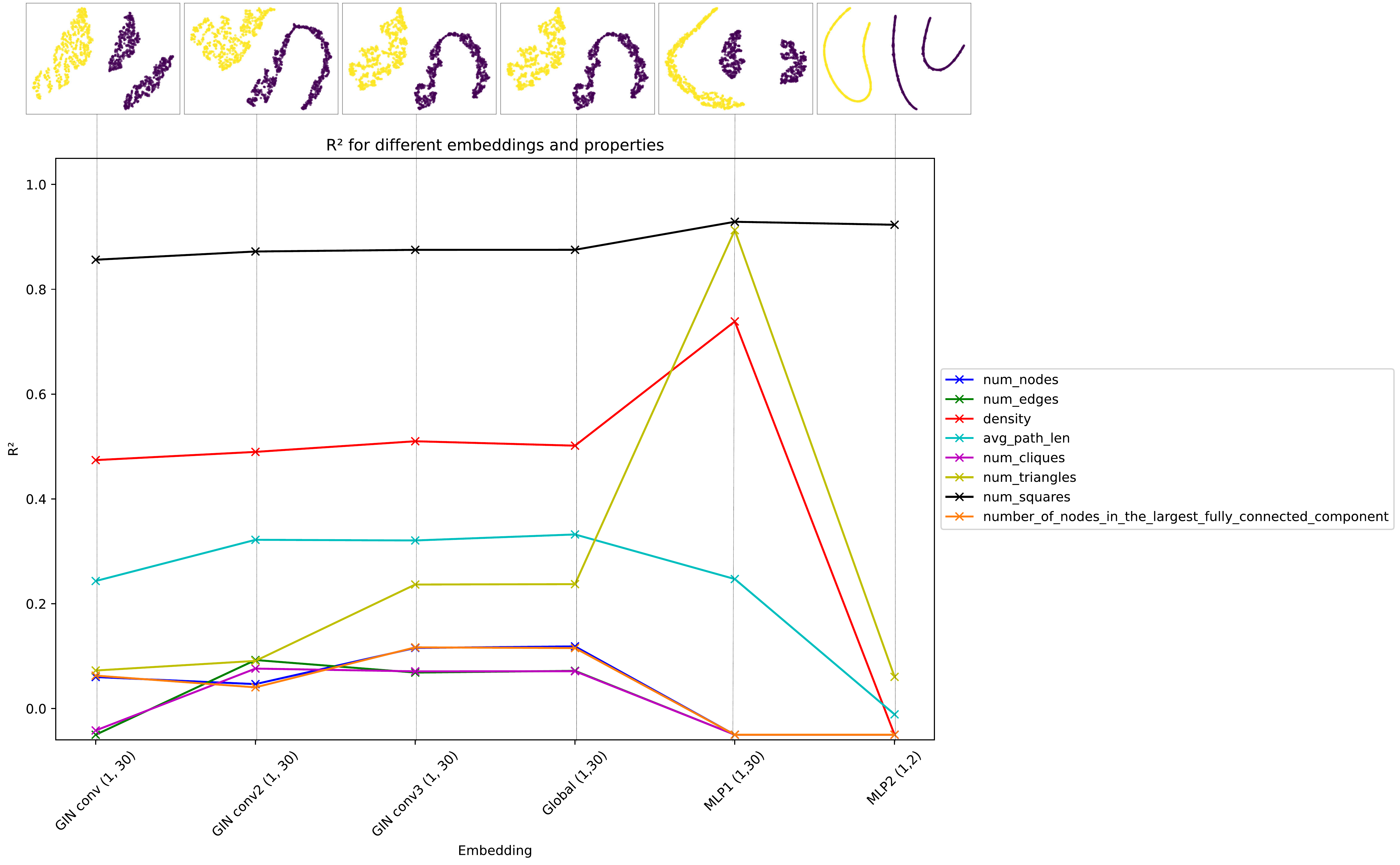}
    \caption{T-SNE visualization across different
layers of our GIN architecture aligned with the probing $R^2$ scores plots with mean-pooled node embeddings (Grid House)}
    \label{fig:plotGINTSNEcontrol_pooled}
\end{figure}

% \begin{figure}
%     \centering
%     \includegraphics[width=1\linewidth]{figures/GIN embeddings.pdf}
%     \caption{T-SNE visualisation across different
% layers of our GIN architecture aligned with the probing $R^2$ scores plots (Grid House)}
%     \label{fig:plotGINTSNEcontrol}
% \end{figure}

% \begin{figure}
%     \centering
%     \includegraphics[width=1\linewidth]{figures/BA_2grid_house_no_isomorphic_GIN4_tsne.pdf}
%     \caption{T-SNE visualisation across different
% layers of our GIN architecture (Grid House)}
%     \label{fig:plotGINTSNEcontrol}
% \end{figure}

% \begin{figure}
%     \centering
%     \includegraphics[width=1\linewidth]{BA_2grid_house_no_isomorphic_GIN4_R2_all_layers.pdf}
%     \caption{Plot of the GIN (control) R2 results across different
% layers probing for graph properties (Grid House)}
%     \label{fig:plotGINcontrol_full}
% \end{figure}

\begin{figure}[h]
    \centering
    \includegraphics[width=1\linewidth]{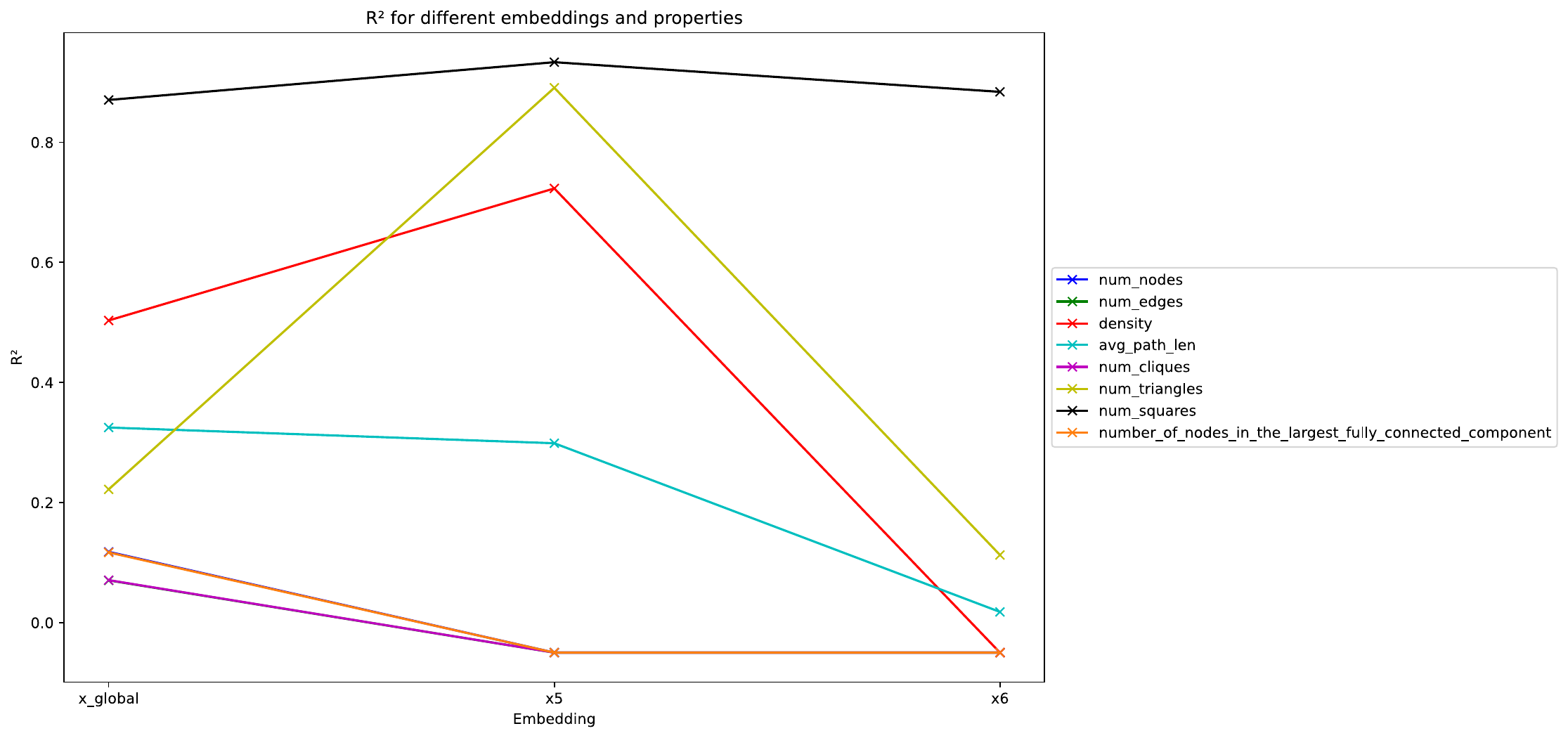}
    \caption{Plot of the GIN (control) $R^2$ results across different
layers probing for graph properties with post pooling layers only, allowing clearer visualization and higher order property interpretation (Grid House)}
    \label{fig:plotGINcontrol}
\end{figure}

\begin{figure}[h]
    \centering
    \includegraphics[width=1\linewidth]{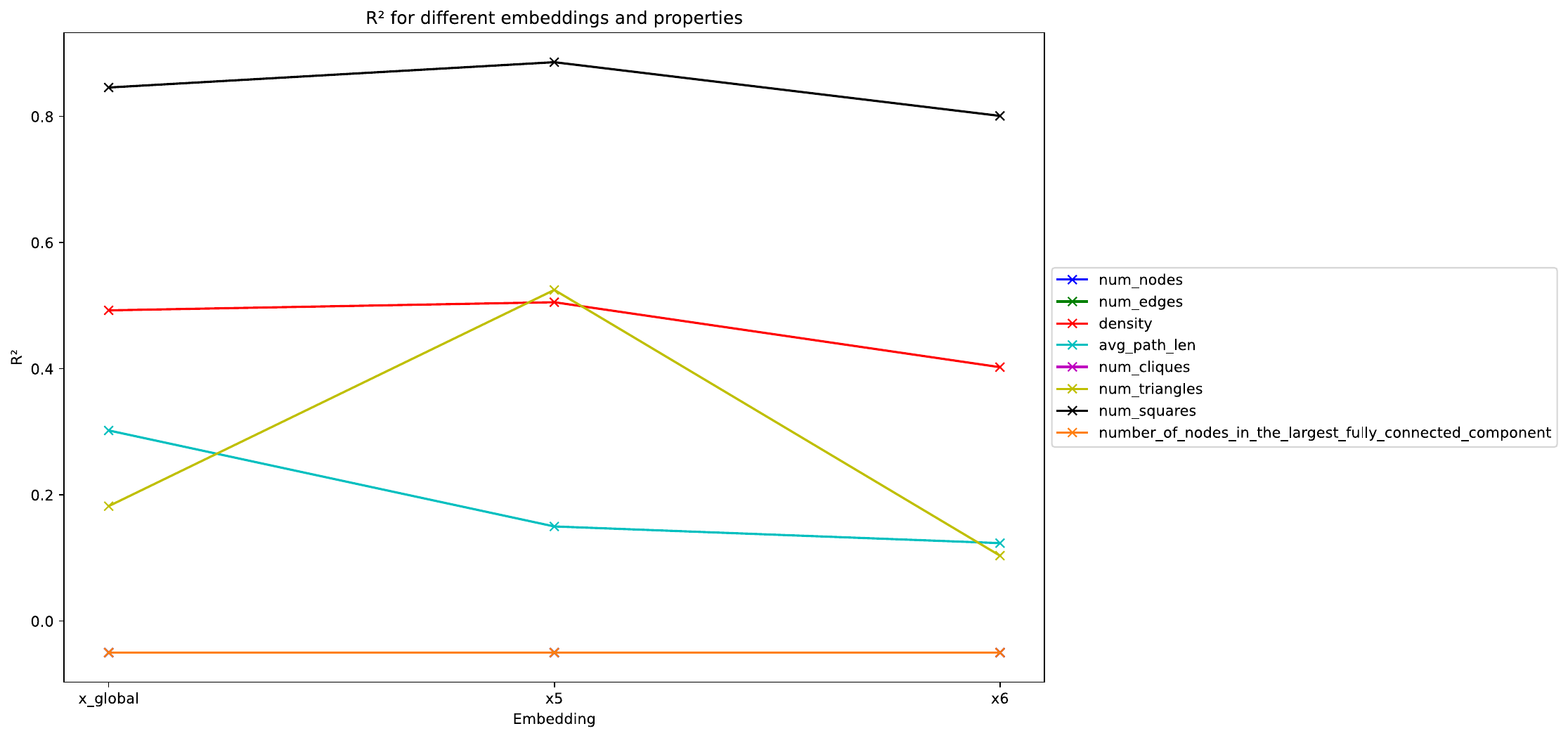}
    \caption{Plot of the GIN ($L_2$) $R^2$ results across different
layers probing for graph properties with post pooling layers only, allowing clearer visualization and higher order property interpretation (Grid House)}
    \label{fig:plotGINL2}
\end{figure}

\begin{figure}[h]
    \centering
    \includegraphics[width=1\linewidth]{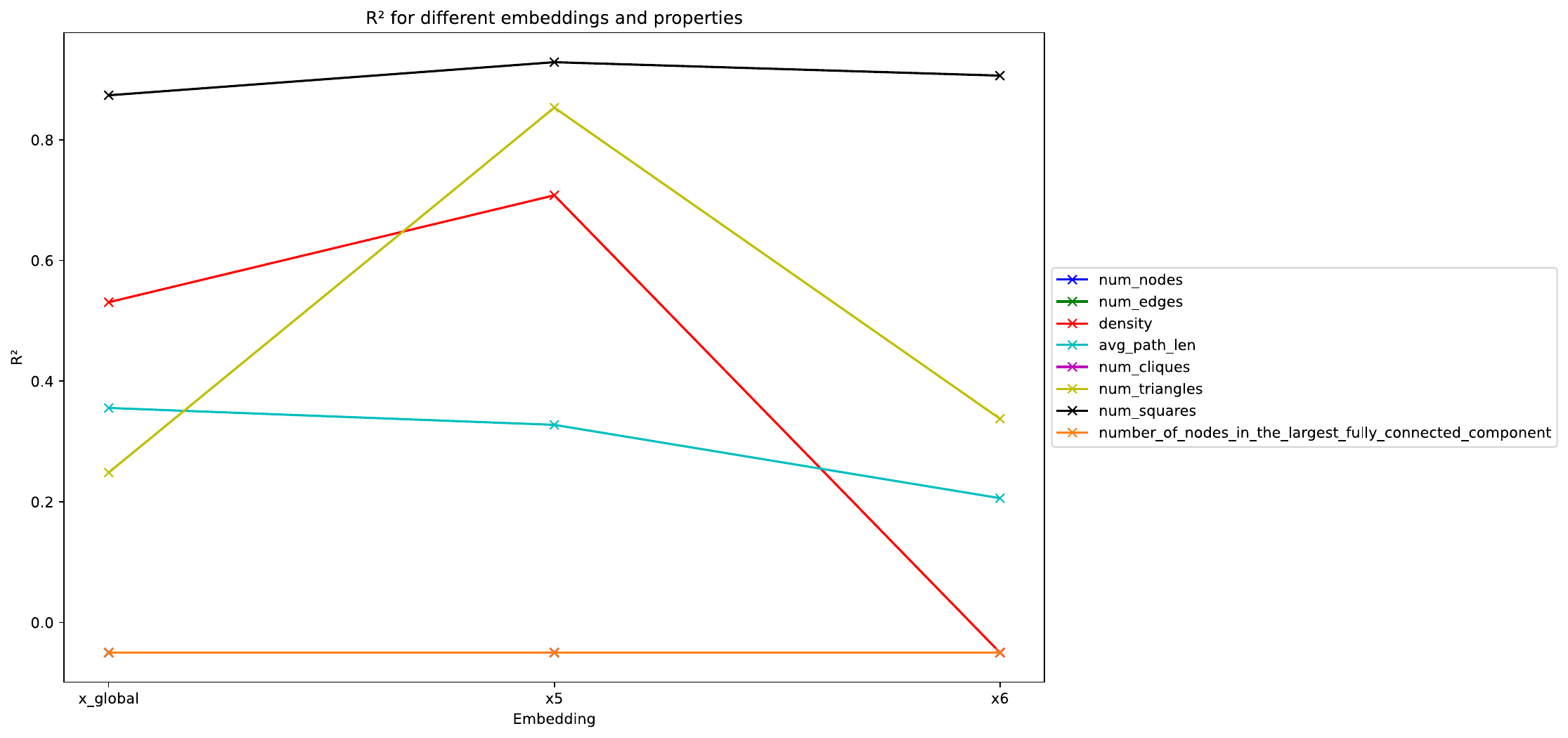}
    \caption{Plot of the GIN (dropout) $R^2$ results across different
layers probing for graph properties with post pooling layers only, allowing clearer visualization and higher order property interpretation (Grid House)}
    \label{fig:plotGINLdropout}
\end{figure}

\newpage
\clearpage
\textbf{GAT}

% \begin{figure}[t!]
%     \centering
%     \includegraphics[width=0.9\linewidth]{figures/TSNE_GAT.pdf}
%     \caption{T-SNE visualization across different
% layers of our GAT architecture aligned with the probing $R^2$ scores plots (Grid House)}
%     \label{fig:plotGATTSNEcontrol}
% \end{figure}

\begin{figure}[h]
    \centering
    \includegraphics[width=0.9\linewidth]{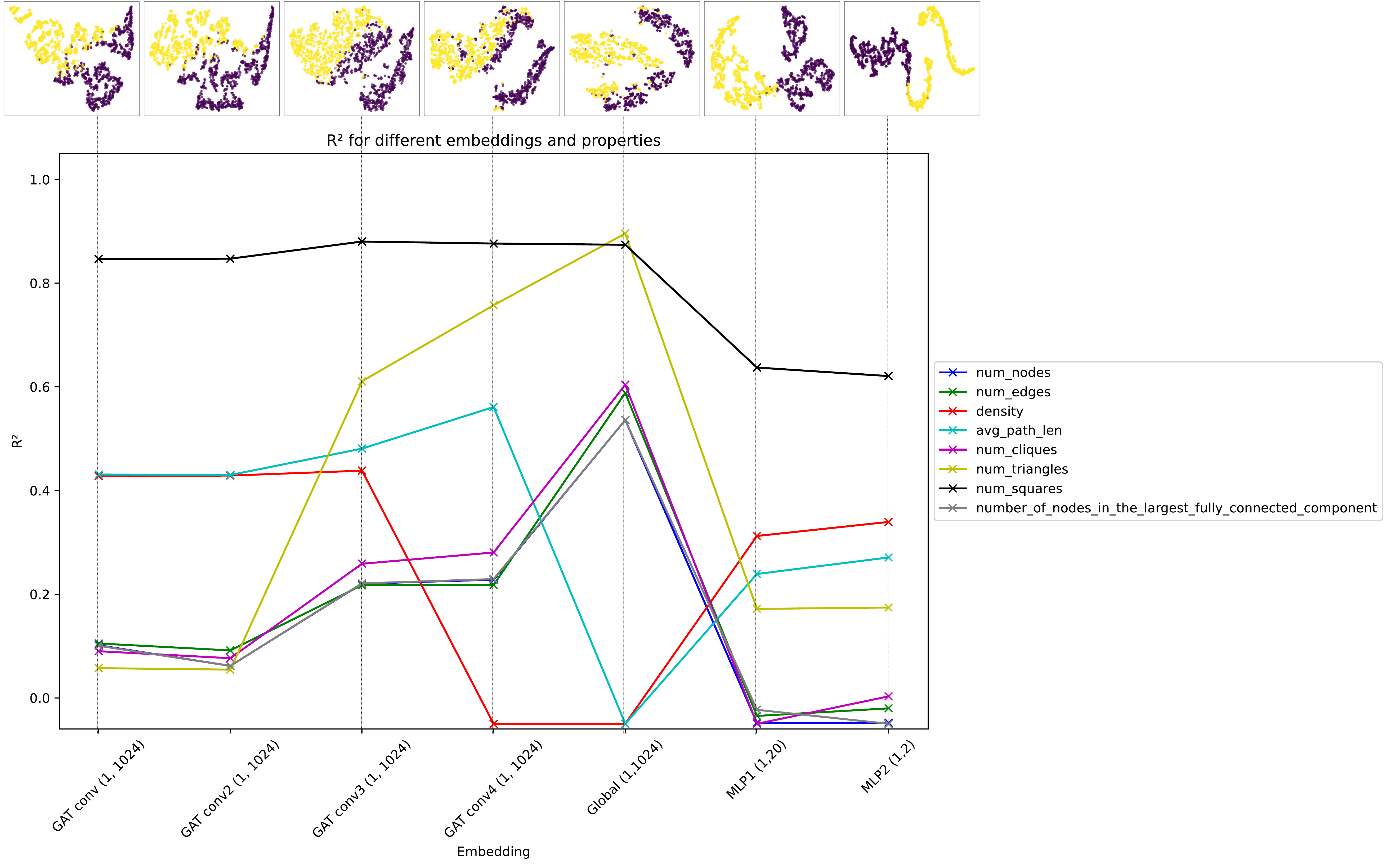}
    \caption{T-SNE visualization across different
layers of our GAT architecture aligned with the probing $R^2$ scores plots with mean-pooled node embeddings (Grid House)}
    \label{fig:plotGATTSNEcontrol_pooled}
\end{figure}

\begin{figure}[h]
    \centering
    \includegraphics[width=1\linewidth]{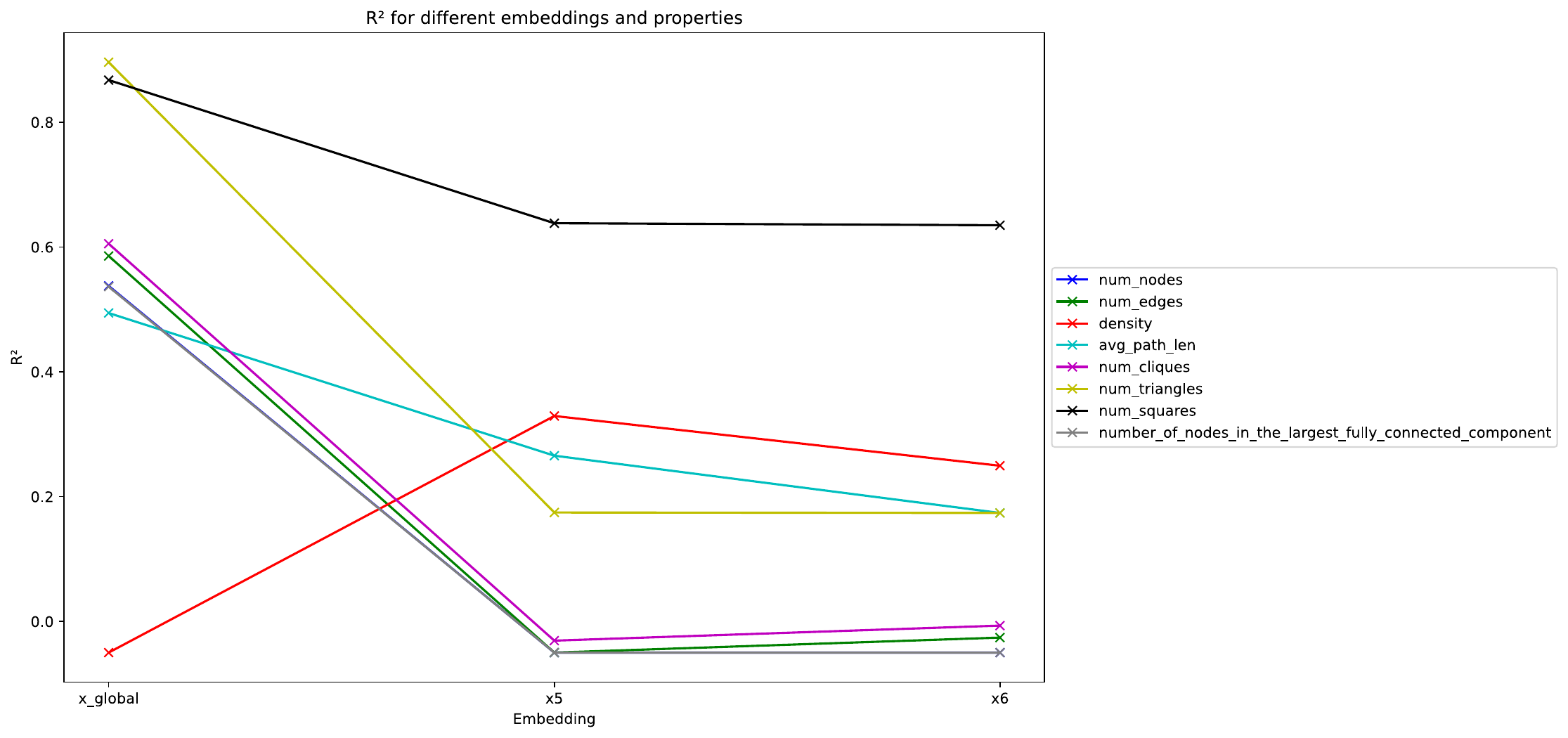}
    \caption{Plot of the GAT $R^2$ results across different
layers probing for graph properties with post pooling layers only, allowing clearer visualization and higher order property interpretation (Grid House)}
    \label{fig:plotGATdropout}
\end{figure}

\clearpage

\subsubsection{Grid House Node properties probing results} \phantom{.} % Adds an invisible character, creating vertical space

Using the probing method developed in the next section, we were not fully able to confirm our initial hypothesis. 

\begin{table}[htpb]
\centering
\caption{Linear Probing $R^2$ Performance Across models for Selected Node Properties (GridHouse Dataset). Best Scores in Bold; Non-convergence indicated by \textemdash}
\begin{tabular}{lccccccccc}
\toprule
\textbf{GCN Layer} & degree & closeness & betweenness & eigenvector & clustering & pagerank \\
\midrule
x1 (GCN) & 0.50 & 0.22 & 0.25 & 0.19 & 0.06 & \textbf{0.56} \\
x2 (GCN) & 0.54 & 0.32 & 0.28 & 0.24 & 0.09 & \textbf{0.57} \\
x3 (GCN) & 0.54 & 0.35 & 0.29 & 0.25 & 0.11 & \textbf{0.57} \\
x4 (GCN) & 0.55 & 0.37 & 0.28 & 0.30 & 0.17 & \textbf{0.57} \\
\bottomrule
\textbf{GIN Layer} & &  &  &  &   \\
\midrule
x1 (GIN) & 0.55 & 0.18 & 0.24 & 0.22 & 0.05 & \textbf{0.56} \\
x2 (GIN) & 0.52 & 0.34 & 0.27 & 0.25 & 0.07 & \textbf{0.54} \\       
\bottomrule
\textbf{GAT Layer} & &  &  &  &   \\
\midrule
Layer 0 & \textbf{0.55} & 0.07 & 0.05 & 0.32 & 0.28 & 0.17 \\
Layer 1 & \textbf{0.52} & 0.48 & 0.08 & 0.31 & 0.30 & 0.14 \\
Layer 2 & 0.47 & \textbf{0.55} & \textemdash & 0.29 & 0.29 & \textemdash \\
Layer 3 & \textbf{0.41} & \textemdash & 0.14 & 0.19 & 0.26 & \textemdash \\
Layer 4 & 0.35 & \textbf{0.50} & 0.12 & 0.21 & 0.23 & \textemdash \\
\bottomrule
\end{tabular}
\label{tab:GridHouseNodeProperties}
\end{table}

In these pre-pooling layers, we first observe the predominance of \textit{page rank} and \textit{node degree} in the early layers and in all the layers of the GCN and the GIN (which has only two of them). When considering the last layers of the GAT (unfortunately we should have have similar architecture with the GIN in order to fully test our hypothesis) it seems that \textit{closeness}, \textit{node degree} and \textit{clustering coefficient} are the most significant. This aligns with our framing of the graph classification task, which is largely driven by the detection of squares and the fact that pre-pooling layers leading to this property detection should affect mostly these three properties. But this does not align with the use of node properties in a graph in order to do graph classification. This still makes a lot of sense. In general, contrary to the graph probing, and to the exception of the node degree, we see that there is not a single property clearly dominating others but that we go towards a combination of different properties just before the graph pooling method. We would have expect the GIN architecture to show similar results with four layers (as we already see an important increase with regard to the closeness between the first and second layer).

\newpage

\clearpage
\section{Clintox dataset}\phantom{.}

\subsection{model}\phantom{.}

\begin{table}[htbp]
    \caption{Performance of Different Models on ClinTox with a 80\%-20\% Random Split. The highest performance is highlighted with boldface. All the performance of methods are reported under their best settings.}
    \label{table:Clintoxperformance}
    \centering
    \begin{tabular}{ccc}
        \hline
        \textbf{Method} & \textbf{ClinTox} \\
        \hline
        GCN & $0.91$  \\
        GAT & $0.92$  \\
        GIN & \textbf{$0.93$}  \\
        \hline
    \end{tabular}
\end{table}

\subsection{Results}

\subsubsection{Graphs properties probing results} \phantom{.} % Adds an invisible character, creating vertical space

\begin{table}[htbp]
\centering
\caption{Linear Probing $R^2$ Performance across the GIN layers for basic graph properties (ClinTox dataset). Best Scores in Bold; Non-convergence indicated by \textemdash (full)}
\label{table:Clintoxbasic_properties}
\begin{tabular}{lcccccc}
\toprule
GIN Layer & \# Nodes & \# Edges & Density & Avg. Path Length & Diameter & Radius \\
\midrule
x1 (GIN) & \textbf{1.00} & \textbf{1.00} & 0.66 & 0.76 & 0.55 & 0.60 \\
x2 (GIN) & \textbf{1.00} & \textbf{1.00} & 0.57 & 0.95 & 0.88** & 0.84 \\
x3 (GIN) & \textbf{1.00} & \textbf{1.00} & 0.62 & \textbf{0.97} & 0.93 & 0.89 \\
x4 (GIN) & \textbf{0.99} & \textbf{0.99} & 0.37 & 0.91 & 0.82 & 0.82 \\
x5 (GIN) & \textbf{0.99} & \textbf{0.99} & 0.29 & 0.90 & 0.82 & 0.82 \\
x\_global & 0.41 & 0.44 & 0.58 & 0.20 & 0.20 & 0.20 \\
x6 (MLP) & 0.40 & 0.44 & 0.58 & 0.19 & 0.19 & 0.19 \\
x7 (MLP) & 0.42 & 0.46 & 0.50 & 0.27 & 0.23 & 0.25 \\
x8 (MLP) & 0.04 & 0.05 & 0.00 & 0.04 & 0.05 & 0.03 \\
\bottomrule
\end{tabular}
\end{table}

\begin{table}[htbp]
\centering
\caption{Linear Probing $R^2$ Performance across the GIN layers for clustering and centrality measures (ClinTox dataset). Best Scores in Bold; Non-convergence indicated by \textemdash (full)}
\label{table:Clintoxclustering_centrality}
\resizebox{\textwidth}{!}{
\begin{tabular}{lcccccc}
\toprule
GIN Layer & Clustering coef. & Transitivity & Assortativity & Avg. clustering & Avg. btw. cent. & PageRank cent. \\
\midrule
x1 (GIN) & \textemdash & \textemdash & 0.32 & \textemdash & \textemdash & 0.18 \\
x2 (GIN) & \textemdash & \textemdash & 0.21 & \textemdash & \textemdash & \textemdash \\
x3 (GIN) & \textemdash & \textemdash & \textemdash & \textemdash & \textemdash & \textemdash \\
x4 (GIN) & \textemdash & \textemdash & \textemdash & \textemdash & \textemdash & \textemdash \\
x5 (GIN) & \textemdash & \textemdash & \textemdash & \textemdash & \textemdash & \textemdash \\
x\_global & \textemdash & \textemdash & 0.25 & \textemdash & 0.48 & \textbf{0.40} \\
x6 (MLP) & \textemdash & \textemdash & \textbf{0.27} & \textemdash & 0.42 & 0.39 \\
x7 (MLP) & \textemdash & \textemdash & \textemdash & \textemdash & \textbf{0.47} & \textemdash \\
x8 (MLP) & \textemdash & \textemdash & \textemdash & \textemdash & 0.06 & \textemdash \\
\bottomrule
\end{tabular}
}
\end{table}

\begin{table}[htbp]
\centering
\caption{Linear Probing $R^2$ Performance across the GIN layers for graph substructures (ClinTox dataset). Best Scores in Bold; Non-convergence indicated by \textemdash (full)}
\label{table:Clintoxgraph_substructures}
\resizebox{\textwidth}{!}{
\begin{tabular}{lcccccc}
\toprule
GIN Layer & \# Cliques & \# Triangles & \# Squares & Largest comp. size & Avg. degree & Graph energy \\
\midrule
x1 (GIN) & 0.99 & \textemdash & 0.00 & 0.99 & 0.53 & \textbf{1.00} \\
x2 (GIN) & \textbf{1.00} & \textemdash & 0.00 & 0.99 & 0.46 & \textbf{1.00} \\
x3 (GIN) & 1.00 & \textemdash & 0.00 & \textbf{0.99} & 0.53 & 1.00 \\
x4 (GIN) & 0.99 & \textemdash & 0.00 & 0.99 & 0.20 & 0.99 \\
x5 (GIN) & 0.99 & \textemdash & 0.00 & 0.99 & \textemdash & 0.99 \\
x\_global & 0.43 & \textemdash & 0.00 & 0.40 & \textbf{0.81} & 0.44 \\
x6 (MLP) & 0.43 & \textemdash & 0.00 & 0.40 & 0.80 & 0.44 \\
x7 (MLP) & 0.46 & \textemdash & 0.00 & 0.42 & 0.75 & 0.46 \\
x8 (MLP) & 0.04 & \textemdash & 0.00 & 0.04 & \textemdash & 0.05 \\
\bottomrule
\end{tabular}
}
\end{table}

\begin{table}[htbp]
\centering
\caption{Linear Probing $R^2$ Performance across the GIN layers for spectral and small-world properties (ClinTox dataset). Best Scores in Bold; Non-convergence indicated by \textemdash (full)}
\label{table:Clintoxspectral_smallworld}
\begin{tabular}{lccccc}
\toprule
GIN Layer & Spectral rad. & Algebraic co. & Small world coef. & Small world idx & Avg. btw. cent. \\
\midrule
x1 (GIN) & 0.70 & 0.78 & \textemdash & \textemdash & \textemdash \\
x2 (GIN) & 0.66 & \textbf{0.80} & \textemdash & \textemdash & \textemdash \\
x3 (GIN) & 0.61 & 0.80 & \textemdash & \textemdash & \textemdash \\
x4 (GIN) & 0.16 & 0.78 & \textemdash & \textemdash & \textemdash \\
x5 (GIN) & \textemdash & 0.69 & \textemdash & \textemdash & \textemdash \\
x\_global & \textbf{0.74} & 0.67 & \textemdash & \textemdash & 0.48 \\
x6 (MLP) & 0.74 & 0.66 & \textemdash & \textemdash & 0.42 \\
x7 (MLP) & 0.71 & 0.56 & \textemdash & \textemdash & \textbf{0.47} \\
x8 (MLP) & 0.07 & 0.02 & \textemdash & \textemdash & 0.06 \\
\bottomrule
\end{tabular}
\end{table}

\clearpage

\subsubsection{Plots} \phantom{.} % Adds an invisible character, creating vertical space

\begin{figure}[htbp]
\centering
\includegraphics[width=1\textwidth]{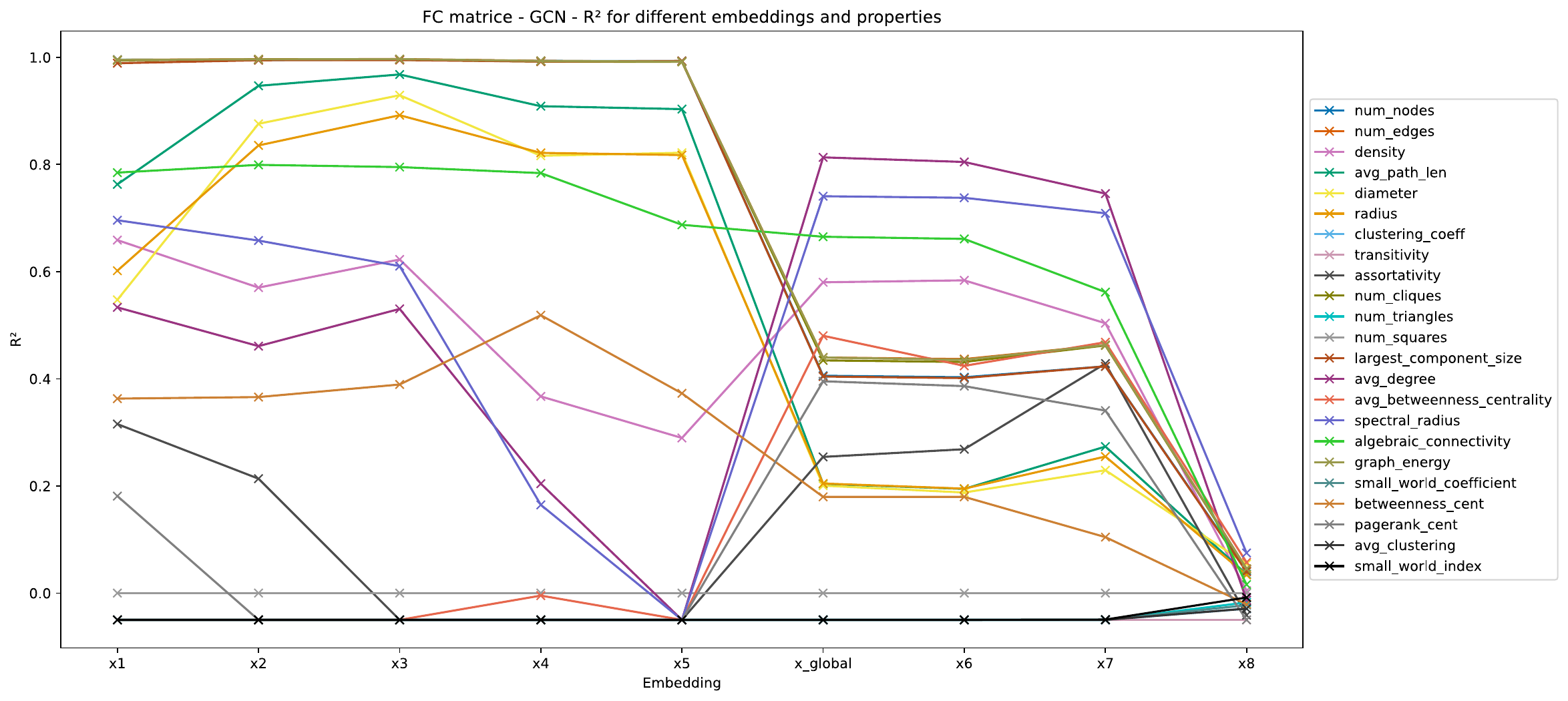}
\caption{Plot of the GIN \(R^2\) results across different layers probing for graph properties. ClinTox dataset (the negative \(R^2\) values have been reduced to -0.05).}
\label{fig:image30}
\end{figure}

\clearpage

\subsubsection{Node properties probing results} \phantom{.} % Adds an invisible character, creating vertical space

\begin{table}[ht]
\small
\centering
\caption{Linear Probing $R^2$ Performance across the GIN layers for various node properties (ClinTox dataset). Best Scores in Bold; Non-convergence indicated by \textemdash}
\begin{tabular}{lcccccc}
\toprule
\textbf{GIN Layer} & degree & closeness & betweenness & eigenvector & clustering & pagerank \\
\midrule
x0 (GIN) & \textbf{0.99} & 0.06 & 0.57 & 0.30 & \textemdash & 0.16 \\
x1 (GIN) & \textbf{0.85} & 0.12 & 0.51 & 0.31 & 0.00 & 0.20 \\
x2 (GIN) & \textbf{0.89} & 0.11 & 0.59 & 0.29 & \textemdash & 0.26 \\
x3 (GIN) & \textbf{0.86} & 0.07 & 0.51 & 0.28 & \textemdash & 0.17 \\
x4 (GIN) & \textbf{0.85} & 0.09 & 0.49 & 0.32 & \textemdash & 0.14 \\
\bottomrule
\end{tabular}
\end{table}

Here again, the very strong presence of the node degree makes a lot of sense when we know this property prepares the aggregation of global properties in the post pooling layers. The interesting thing is the non negligible presence of the betweenness centrality in all the layers which suggests that the betweenness centrality of atoms is important in the aggregation of global molecule properties that help predict the toxicity of a molecule. This property is more than the closeness or the clustering coefficient. The irreplaceable nature of some atoms in the molecular graph, which is literally the meaning of having a high betweenness centrality, is an important feature which makes these atoms targets to be part of higher order molecular schemes and patterns.

\begin{figure}[htbp]
\centering
\includegraphics[width=1\textwidth]{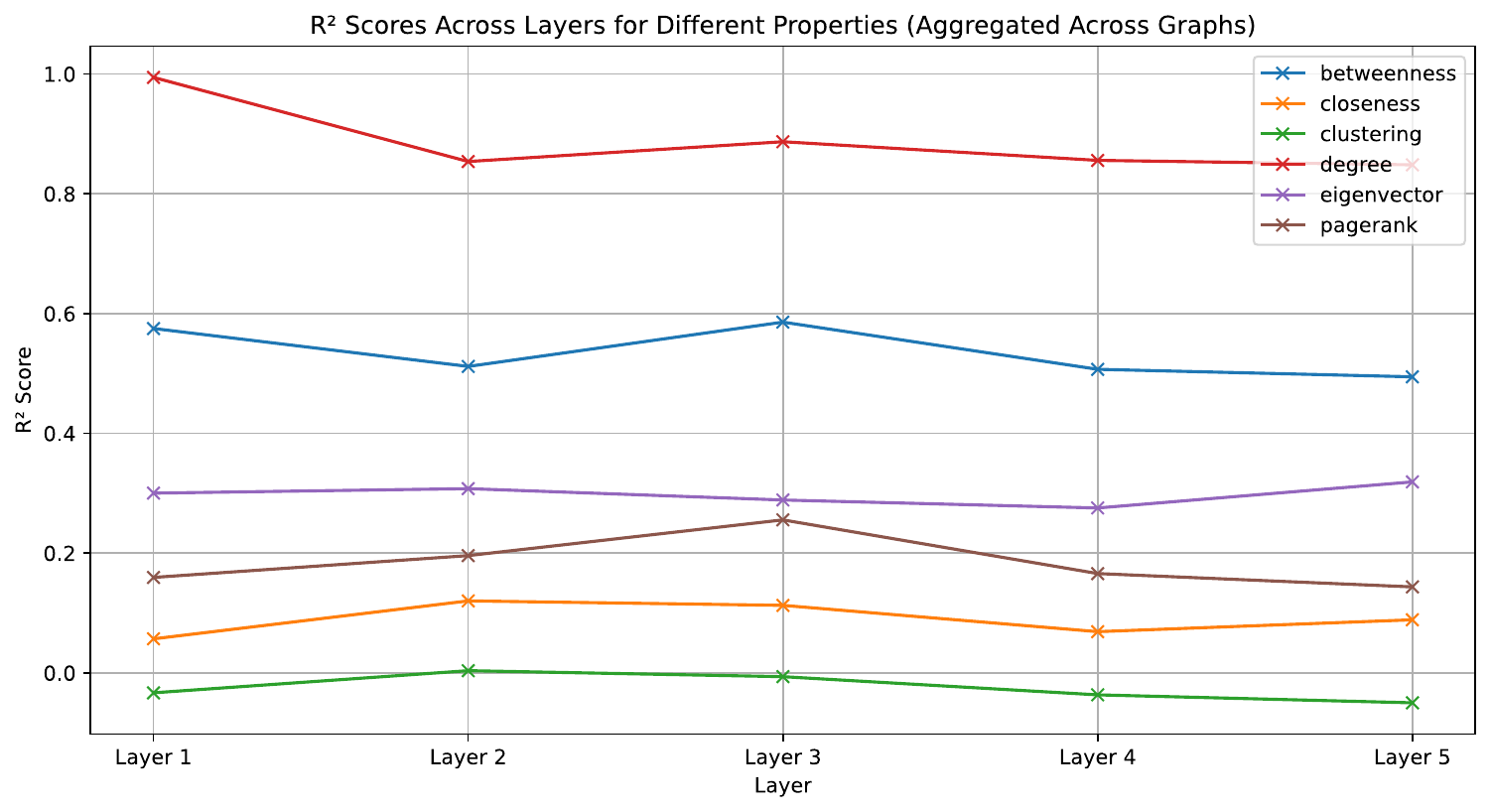}
\caption{Plot of the GIN \(R^2\) results across different layers probing for node properties. ClinTox dataset (the negative \(R^2\) values have been reduced to -0.05). (full results)}
\label{fig:image31}
\end{figure}

\end{document}